\documentclass[letterpaper, 10 pt, conference]{ieeeconf}  

\IEEEoverridecommandlockouts                              

\usepackage{graphics} 
\usepackage{subfigure}
\usepackage{epsfig} 
\usepackage{mathptmx} 
\usepackage{times} 
\usepackage{amsmath} 
\usepackage{amssymb}  
\usepackage{booktabs}	%
\usepackage{balance}

\title{\LARGE \bf
LINS: A Lidar-Inertial State Estimator for Robust and Efficient Navigation
}

\author{Chao Qin$^{1}$,  Haoyang Ye$^{1}$, Christian E. Pranata$^{1}$, Jun Han, Shuyang Zhang$^{1}$, and Ming Liu$^{1}$
\thanks{$^{1}$The authors are with the RAM lab, the Hong Kong University of Science and Technology, Kowloon, Hong Kong {\tt\small cscharlesqin@gmail.com, hy.ye@connect.ust.hk, shuyang.zhang1995@gmail.com, christianedwinp@yahoo.com, eelium@ust.hk}}%
}

\begin{document}

\maketitle
\thispagestyle{empty}
\pagestyle{empty}

\begin{abstract}
We present LINS, a lightweight lidar-inertial state estimator, for real-time ego-motion estimation. The proposed method enables robust and efficient navigation for ground vehicles in challenging environments, such as feature-less scenes, via fusing a 6-axis IMU and a 3D lidar in a tightly-coupled scheme. An iterated error-state Kalman filter (ESKF) is designed to correct the estimated state recursively by generating new feature correspondences in each iteration, and to keep the system computationally tractable. Moreover, we use a robocentric formulation that represents the state in a moving local frame in order to prevent filter divergence in a long run. To validate robustness and generalizability, extensive experiments are performed in various scenarios. Experimental results indicate that LINS offers comparable performance with the state-of-the-art lidar-inertial odometry in terms of stability and accuracy and has order-of-magnitude improvement in speed.
\end{abstract}

\section{INTRODUCTION}
Ego-motion estimation is a fundamental prerequisite to enable most mobile robotic applications---poor real-time capability and failure of the algorithm can quickly lead to damage of the hardware and its surroundings. To this end, active sensors, such as lidars, are proposed to fulfill this task, which widely known as simultaneous localization and mapping (SLAM). Some of the key advantages of a typical 3D lidar include (i) wide horizontal field-of-view (FOV) \cite{velas2016collar} and (ii) invariance to ambient lighting conditions \cite{barfoot2016into}. However, the lidar-based navigation system is sensitive to surroundings. Furthermore, the motion distortion \cite{anderson2013ransac} and sparse nature of point clouds \cite{behley2018efficient} make it even worse in some challenging scenarios (e.g. a wide and open area). 

Recent research has demonstrated that deficiencies of a stand-alone lidar can be compensated by fusing an IMU. An IMU, unlike lidar, is insensitive to surroundings. It provides accurate short-term motion constraints and generally works at a high frequency (e.g., 100 Hz-500 Hz). These features can help the lidar navigation system to recover point clouds from highly dynamic motion distortion and thus increase accuracy. However, the state-of-the-art lidar-inertial odometry (LIO) \cite{ye2019tightly} which is based on graph optimization cannot be directly applied to real-time navigation due to high computational expense; for a single scan, it takes more than 100 millisecond to compute the lidar-inertial odometry and even more time to maintain a map.

\begin{figure}[t]
\centering
\framebox{\includegraphics[width=0.45\textwidth]{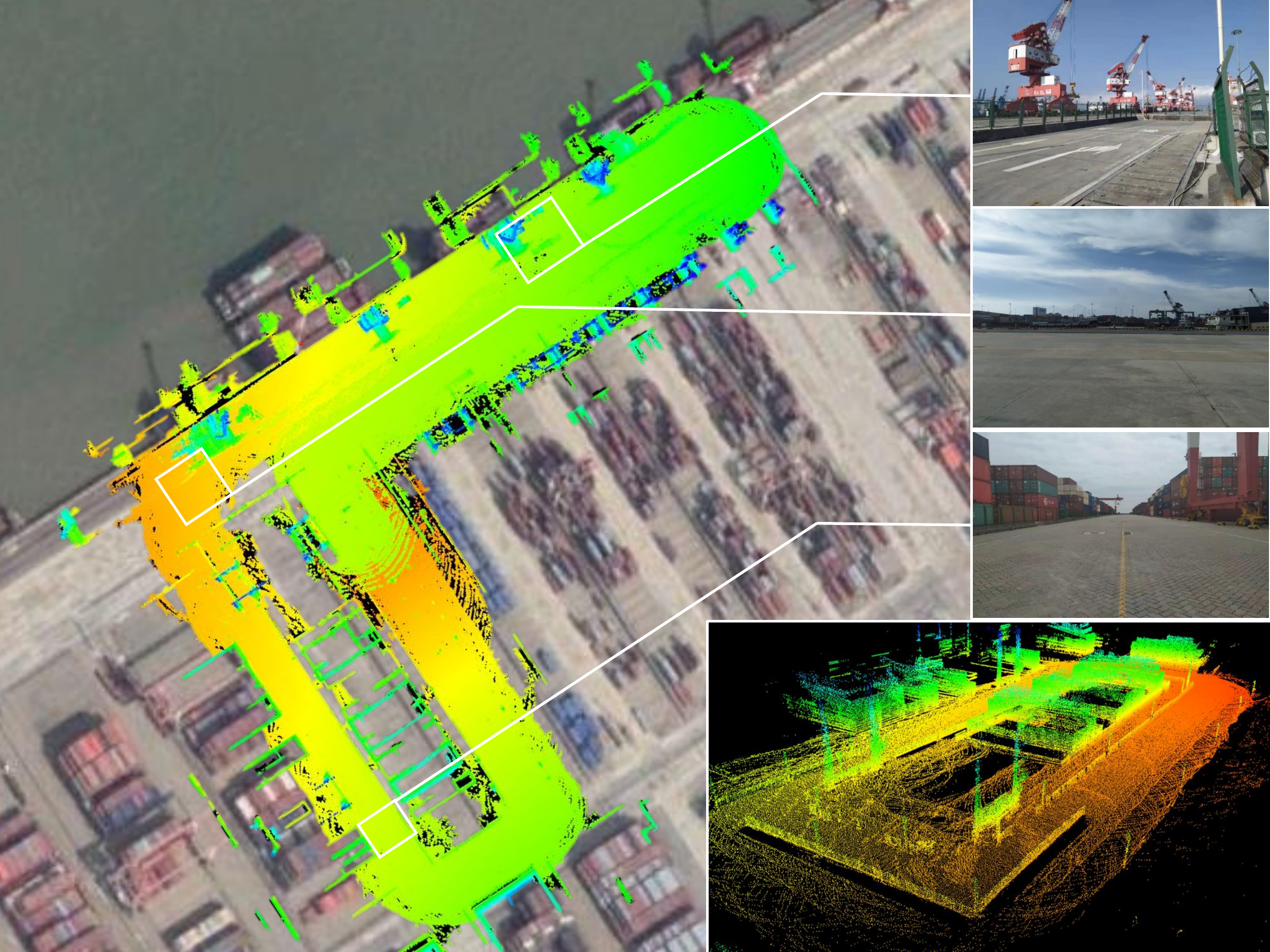}}
	\caption{3D Map built by LINS with a Velodyne VLP-16 and an Xsens MTi-G-710 IMU in a port in Guangdong. We can observe the good alignment of the resulting map with Google Map, even in some feature-less environments as shown in images on the right.}
	\label{fig_pcl_in_googlemap}
\end{figure}

In this paper, we propose LINS, a lightweight lidar-inertial state estimator for real-time navigation of unmanned-ground-vehicles (UGVs). An iterated error-state Kalman filter (ESKF) is designed to ensure both accuracy and efficiency. To achieve long-term stability, we introduce a robocentric formulation of the state in which the local frame of reference is shifted at every lidar time-step, and the relative pose estimate between two consecutive local frames is used for updating the global pose estimate. The main contributions of our work are as follows:
\begin{itemize}
	\item A tightly-coupled lidar-inertial odometry algorithm, which is faster than our previous work \cite{ye2019tightly} by an order of magnitude, is proposed.
	\item We present a robocentric iterated ESKF, which is verified in various challenging scenarios and shows superior performance over the state of the art. 
	\item The source code is available online\footnote{https://github.com/ChaoqinRobotics}. To the best of our knowledge, LINS is the first tightly-coupled LIO that solves the 6 DOF ego-motion via iterated Kalman filtering.
\end{itemize}

The remaining paper is organized as follows. In Sect. \ref{Related Work}, we discuss relevant literature. We give an overview of the complete system pipeline in Sect. \ref{Lidar-Inertial Odometry and Mapping}. The experimental results are illustrated in Sect. \ref{Experiments}, followed by a conclusion in Sect. \ref{Conclusion}.

\begin{figure*}[t]
	\centering
	\framebox{\includegraphics[width=0.9\textwidth]{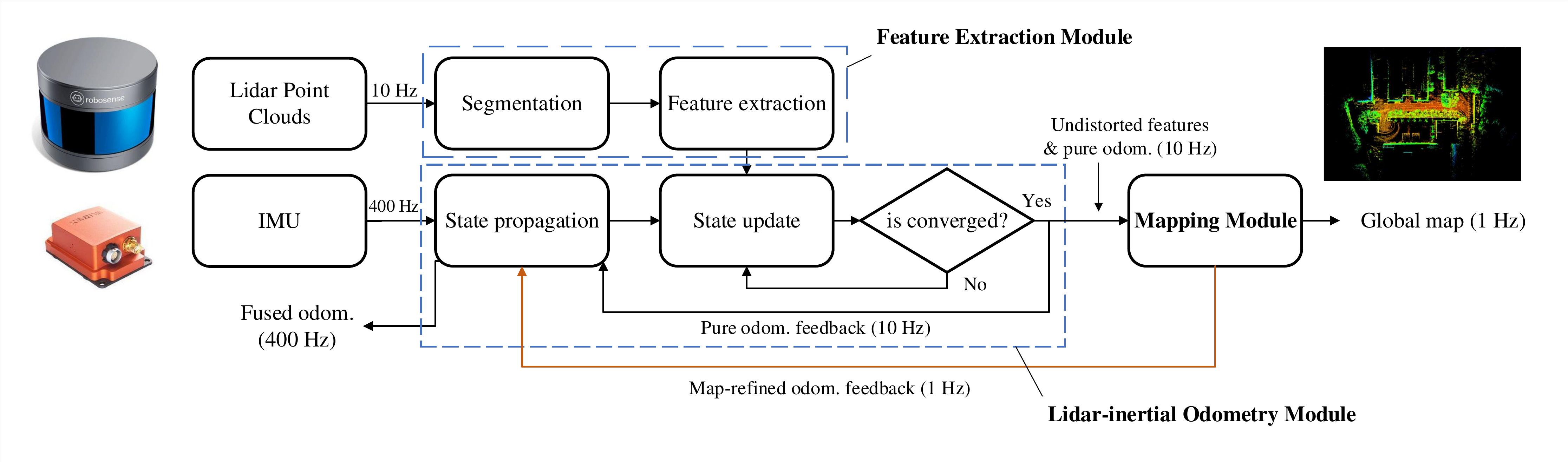}}
	\caption{Pipeline of the LINS system. The lidar-inertial odometry module, which consists of state propagation and update submodule, performs iterated Kalman filtering using IMU measurements and point cloud features extracted from the feature extraction module. The mapping module outputs a refined pose estimate along with a global 3D map. The refined pose estimate is combined with IMU measurements to generate high-output-rate results. Note that the focus of this work is the lidar-inertial odometry module.}
	\label{fig_lins_flow_chart}
\end{figure*}

\section{Related Work}\label{Related Work}
There are hundreds of works on lidar-related odometry in the literature. We restrict our attention to related work on the 6 DOF ego-motion estimators and relevant fusion algorithms, which are categorized into the loosely-coupled and the tightly-coupled. 

\subsection{Lidar-only Methods} 
Many lidar-only approaches are variations of the well-known iterative closest point (ICP) scan matching method which is based on scan-to-scan registration. \cite{rusinkiewicz2001efficient, pomerleau2015review} had surveyed efficient variants of ICP. For real-time application, \cite{zhang2014loam} established LOAM which sequentially registers extracted edge and planar features to an incrementally built global map. \cite{shan2018lego} proposed LeGO which adapts the original LOAM to UGV application. By applying ground plane extraction and point cloud segmentation, LeGO filtered out unreliable features and showed great stability in areas covered with noisy objects, i.e., grass and trees. \cite{hess2016real} provided an efficient loop closure mechanism to enable large-scale mapping in real time.

\subsection{Loosely-coupled Lidar-IMU Fusion} 
Loosely-coupled methods deal with two sensors separately to infer their motion constraints which are fused later (e.g., \cite{tang2015lidar, lynen2013robust}). IMU-aided LOAM \cite{zhang2014loam} took the orientation and translation calculated by the IMU as priors for optimization. \cite{zhen2017robust} combined the IMU measurements with pose estimates obtained from a lidar-based Gaussian particle filter and a pre-built map. In general, loosely-coupled fusion is computationally efficient \cite{li2013real}, but the decoupling of lidar and inertial constraints results in information loss \cite{huai2018robocentric}.

\subsection{Tightly-coupled Lidar-IMU Fusion} 
Tightly-coupled approaches directly fuse lidar and inertial measurements through joint-optimization, which can be categorized into optimization based \cite{qin2018vins, leutenegger2013keyframe} and extended Kalman filter (EKF) based \cite{huang2011observability, huai2018robocentric}. \cite{park2018elastic} performed local trajectory optimization via minimizing constraints from the IMU and lidar together. \cite{geneva2018lips} presented LIPS which leveraged graph optimization over the inertial pre-integration constraints \cite{forster2015imu} and plane constraints from a lidar. \cite{ye2019tightly} established LIO-mapping (for brevity, termed LIOM in the followings) which is also based on graph optimization but with a novel rotation-constrained mapping method to optimize the final poses and maps. However, constraint construction and batch optimization in a local map window are too time-consuming for real-time application. \cite{hesch2010laser} introduced a lidar-aided inertial EKF based on a 2D lidar. But its application scenarios were limited to indoor environments because it required that all surrounding planes are in an orthogonal structure. 

It is well known that EKF is vulnerable to linearization errors which may cause poor performance and even lead to divergence \cite{huang2013quadratic, huang2007convergence}. This shortcoming becomes salient when it involves lidar-observed scan-to-scan constraints, which is deemed highly nonlinear if the initial pose is incorrect and causes wrong feature-matching results. To eliminate errors caused by wrong matching, we presented an iterated Kalman filtering \cite{bell1993iterated} which can repeatedly find better matchings in each iteration. Besides, we adopted an error-state representation to guarantee linearization validity \cite{sola2017quaternion}. This characteristic distinguishes our method from the iterated extented Kalman filter \cite{barfoot2017state}.

\section{Lidar-Inertial Odometry and Mapping}\label{Lidar-Inertial Odometry and Mapping}
\subsection{System Overview}
Consider an UGV equipped with an IMU and a 3D lidar. Our goal is to estimate its 6 DOF ego-motion and to establish a global map simultaneously, as shown in Fig. \ref{fig_pcl_in_googlemap}. An overview of the system framework is depicted in Fig. \ref{fig_lins_flow_chart}. The overall system consists of three major modules: feature extraction, LIO, and mapping. (i) The feature extraction module aims at extracting stable features from raw point clouds. (ii) The LIO module, which consists of propagation and update submodules, carries out iterated Kalman filtering and outputs an initial odometry along with undistorted features. (iii) The mapping module refines the initial odometry by the global map and outputs a new odometry, followed by updating the map using new features. Due to the space issue, we only focus on the odometry module. We refer the reader to \cite{zhang2014loam, shan2018lego} for detailed procedures of feature extraction and mapping.

\subsection{Feature Extraction}\label{Feature Extraction}
This module inputs the raw point cloud and outputs a group of edge features, $\mathbb{F}_e$, and a group of planar features, $\mathbb{F}_p$. Readers can see \cite{shan2018lego, zhang2014loam} for detailed implementations.

\subsection{Lidar-Inertial Odometry with Iterated ESKF}\label{Lidar-Inertial Odometry with IESKF}
The LIO module uses IMU measurements and features extracted in two consecutive scans to estimate the relative transformation of the vehicle. We use a robocentric formulation to build the iterated ESKF because it prevents large linearization errors caused by ever-growing uncertainty \cite{civera20101, bloesch2017iterated}. Let $\mathcal{F}_{w}$ represent the fixed world frame, $\mathcal{F}_{b_k}$ represent the IMU-affixed frame at $k$ lidar time-step, and $\mathcal{F}_{l_k}$ represent the lidar frame at $k$ lidar time-step. Note that, in our work, the local frame is always set as the IMU-affixed frame at previous lidar time-step.

\subsubsection{State Definitions}
Let $\mathbf{x}_{w}^{b_k}$ denote the location of $\mathcal{F}_{w}$ w.r.t. $\mathcal{F}_{b_k}$. Let $\mathbf{x}_{b_{k+1}}^{b_k}$ denote the local state that describes the relative transformation from $\mathcal{F}_{b_{k+1}}$ to $\mathcal{F}_{b_k}$:
\begin{align}
\mathbf{x}_{w}^{b_k}&:=\left[\mathbf{p}_{w}^{b_k},\mathbf{q}_{w}^{b_k}\right],\\
\mathbf{x}_{b_{k+1}}^{b_{k}}&:=\left[\mathbf{p}_{b_{k+1}}^{b_{k}},\mathbf{v}_{b_{k+1}}^{b_{k}},\mathbf{q}_{b_{k+1}}^{b_{k}},\mathbf{b}_{a},\mathbf{b}_{g},\mathbf{g}^{b_{k}}\right],
\end{align}
where $\mathbf{p}_{w}^{b_k}$ is the position of $\mathcal{F}_{w}$ w.r.t. $\mathcal{F}_{b_k}$ and $\mathbf{q}_{w}^{b_k}$ is the unit quaternion describing the rotation from $\mathcal{F}_{w}$ to $\mathcal{F}_{b_k}$. $\mathbf{p}_{b_{k+1}}^{b_{k}}$ and $\mathbf{q}_{b_{k+1}}^{b_{k}}$ represent the translation and rotation from $\mathcal{F}_{b_{k+1}}$ to $\mathcal{F}_{b_k}$. $\mathbf{v}_{b_{k+1}}^{b_{k}}$ is the velocity w.r.t. $\mathcal{F}_{b_k}$. $\mathbf{b}_{a}$ is the acceleration bias and $\mathbf{b}_{g}$ the gyroscope bias. It is important to note that the local gravity, $\mathbf{g}^{b_{k}}$ (represented in $\mathcal{F}_{b_k}$), is also part of the local state.

For having good properties in state estimation \cite{madyastha2011extended}, an \textit{error-state} representation is used to solve $\mathbf{x}_{b_{k+1}}^{b_{k}}$. We denote an error term with $\delta$ and define the error vector of $\mathbf{x}_{b_{k+1}}^{b_k}$ as
\begin{equation}
\mathbf{\delta x}:=\left[\mathbf{\delta p},\mathbf{\delta v},\delta\boldsymbol{\theta},\mathbf{\delta b}_{a},\mathbf{\delta b}_{g},\delta\mathbf{g}\right],
\end{equation}
where $\delta\boldsymbol{\theta}$ is a 3 DOF error angle. 

According to the ESKF traditions, once $\mathbf{\delta x}$ is solved, we can obtain the final $\mathbf{x}_{b_{k+1}}^{b_{k}}$ by injecting $\mathbf{\delta x}$ into the state prior of $\mathbf{x}_{b_{k+1}}^{b_{k}}$, $^{-\!}\mathbf{x}_{b_{k+1}}^{b_k}$. This is conducted via a boxplus operator $\boxplus$ which is defined as:
\begin{equation} \label{Box plus}
\mathbf{x}_{b_{k+1}}^{b_{k}}=^{-\!}\mathbf{x}_{b_{k+1}}^{b_{k}}\boxplus\delta\mathbf{x}=\left[\begin{array}{c}
^{-\!}\mathbf{p}_{b_{k+1}}^{b_{k}}+\delta\mathbf{p}\\
^{-\!}\mathbf{v}_{b_{k+1}}^{b_{k}}+\delta\mathbf{v}\\
^{-\!}\mathbf{q}_{b_{k+1}}^{b_{k}}\otimes\exp(\delta\boldsymbol{\theta})\\
^{-\!}\mathbf{b}_{a}+\delta\mathbf{b}_{a}\\
^{-\!}\mathbf{b}_{g}+\delta\mathbf{b}_{g}\\
^{-\!}\mathbf{g}^{b_{k}}+\delta\mathbf{g}
\end{array}\right],
\end{equation}
where $\otimes$ denotes the quaternion product and $\exp:\mathbb{R}^{3}\rightarrow SO(3)$ maps the angle vector to quaternion rotation \cite{bloesch2016primer}.

\subsubsection{Propagation}
In this step, we propagate the error state, $\mathbf{\delta x}$, the error-state covariance matrix, $\mathbf{P}_{k}$, and the state prior, $^{-\!}\mathbf{x}_{b_{k+1}}^{b_k}$, if a new IMU measurement arrives. The linearized continuous-time model \cite{shin2005estimation} for the IMU error state is written as
\begin{equation}\label{IMU model}
\delta\dot{\mathbf{x}}(t)=\mathbf{F}_t\delta\mathbf{x}(t)+\mathbf{G}_t\mathbf{w},
\end{equation}
where $\mathbf{w}=[\mathbf{n}_{a}^T,\mathbf{n}_{g}^T,\mathbf{n}_{b_{a}}^T,\mathbf{n}_{b_{g}}^T]^{T}
$ is the Gaussian noise vector (whose definitions are the same as \cite{qin2018vins}). $\mathbf{F}_t$ is the error-state transition matrix and $\mathbf{G}_t$ is the noise Jacobian at time $t$:
\begin{equation}
\mathbf{F}_{t}=\left[\begin{array}{cccccc}
0 & \mathbf{I} & 0 & 0 & 0 & 0\\
0 & 0 & -\mathbf{R}_{t}^{b_{k}}[\hat{\mathbf{a}}_{t}]_{\times} & -\mathbf{R}_{t}^{b_{k}} & 0 & 0\\
0 & 0 & -[\hat{\boldsymbol{\omega}}_{t}]_{\times} & 0 & -\mathbf{I}_{3} & -\mathbf{I}_{3}\\
0 & 0 & 0 & 0 & 0 & 0\\
0 & 0 & 0 & 0 & 0 & 0\\
0 & 0 & 0 & 0 & 0 & 0
\end{array}\right],
\end{equation}
\begin{equation}
\mathbf{G}_{t}=\left[\begin{array}{cccc}
0 & 0 & 0 & 0\\
-\mathbf{R}_{t}^{b_{k}} & 0 & 0 & 0\\
0 & -\mathbf{I}_{3} & 0 & 0\\
0 & 0 & \mathbf{I}_{3} & 0\\
0 & 0 & 0 & \mathbf{I}_{3}\\
0 & 0 & 0 & 0
\end{array}\right],
\end{equation}
where $[\cdot]_{\times}\in\mathbb{R}^{3\times3}$ transfers a 3D vector to its skew-symmetric matrix. $\mathbf{I}_3\in\mathbb{R}^{3\times3}$ is the identity matrix and $\mathbf{R}_{t}^{b_{k}}$ is the rotation matrix from the IMU-affixed frame at time $t$ to $\mathcal{F}_{b_k}$. $\hat{\mathbf{a}}_{t}$ and $\hat{\boldsymbol{\omega}}_{t}$ are acceleration and angular rate at time $t$, respectively, and they are calculated by removing the biases and the gravity effect from the raw accelerometer measurement,  $\mathbf{a}_{m_t}$, and gyroscope measurements, $\boldsymbol{\omega}_{m_t}$, as
\begin{align}
\hat{\mathbf{a}}_{t}&=\mathbf{a}_{m_{t}}-\mathbf{b}_{a},\\
\hat{\boldsymbol{\omega}}_{t}&=\boldsymbol{\omega}_{m_{t}}-\mathbf{b}_{g}.
\end{align}

Discretizing Equation (\ref{IMU model}) yields following propagation equations:
\begin{equation}
\delta\mathbf{x}_{t_{\tau}}=(\mathbf{I}+\mathbf{F}_{t_{\tau}}\Delta t)\delta\mathbf{x}_{t_{\tau-1}},
\end{equation}
\begin{equation}
\mathbf{P}_{\!t_{\tau}\!}\!=\!(\mathbf{I}+\mathbf{F}_{\!t_{\tau}\!}\Delta t)\mathbf{P}_{\!t_{\tau\!-\!1}\!}(\mathbf{I}+\mathbf{F}_{\!t_{\tau}\!}\Delta t)^{T}\!+\!(\mathbf{G}_{\!t_{\tau}\!}\Delta t)\mathbf{Q}(\mathbf{G}_{\!t_{\tau}\!}\Delta t)^{T},
\end{equation}
where $\Delta t=t_{\tau}-t_{\tau-1}$. $t_{\tau}$ and $t_{\tau-1}$ are consecutive IMU time-steps. $\mathbf{Q}$ expresses the covariance matrix of $\mathbf{w}$, which is computed off-line during sensor calibration. 

To predict $^{-\!}\mathbf{x}_{b_{k+1}}^{b_k}$, the discrete-time propagation model for the robocentric state is required. Readers can refer to \cite{huai2018robocentric, sola2017quaternion} for details in integrating IMU measurements.

\subsubsection{Update}
We now present the iterated update scheme, which is the primary contribution of this work.

In iterated Kalman filtering, the state update can be linked to an optimization problem \cite{pomerleau2015review, bloesch2017iterated} considering the deviation from the prior $^{-}\mathbf{x}_{b_{k+1}}^{b_{k}}$ and the residual functions, $f(\cdot)$, (i.e., innovation\footnote{Note that innovation is the difference between the actual an expected measurements \cite{barfoot2017state}}) derived from the measurement model:
\begin{equation}\label{Cost Function}
\underset{\delta\mathbf{x}}{\min}\|\delta\mathbf{x}\|_{(\mathbf{P}_{k})^{-1}}+\| f(^{-}\mathbf{x}_{b_{k+1}}^{b_{k}}\boxplus\delta\mathbf{x})\|_{(\mathbf{J}_{k}\mathbf{M}_{k}\mathbf{J}_{k}^{T})^{-1}},
\end{equation}
where $\|\cdot\|$ denotes the Mahalanobis norm. $\mathbf{J}_{k}$ is the Jacobian of $f(\cdot)$ w.r.t. the measurement noise and $\mathbf{M}_{k}$ is the covariance matrix of the measurement noise. The output of $f(\cdot)$ is actually a stacked residual vector calculated from point-edge or point-plane pairs. Given $\mathbf{x}_{b_{k+1}}^{b_{k}}$, the error term in $f(\cdot)$ that corresponds to $\mathbf{p}_{i}^{l_{k+1}}$, the $i$ th feature point which is represented in $\mathcal{F}_{l_{k+1}}$, can be described as:
\begin{equation}
f_{i}(\mathbf{x}_{b_{k+1}}^{b_{k}}\!)\!=\!\begin{cases}\label{Edge}
\!\frac{|(\hat{\mathbf{p}}_{i}^{l_{k}}-\mathbf{p}_{a}^{l_{k}})\times(\hat{\mathbf{p}}_{i}^{l_{k}}-\mathbf{p}_{b}^{l_{k}})|}{|\mathbf{p}_{a}^{l_{k}}-\mathbf{p}_{b}^{l_{k}}|}\! & \text{if}~\mathbf{p}_{i}^{l_{k\!+\!1}}\!\in\!\mathbb{F}_{e}\!  \\
\!\frac{|(\hat{\mathbf{p}}_{i}^{l_{\!k\!}}\!-\!\mathbf{p}_{a}^{l_{\!k\!}})^{\!T\!}\!((\mathbf{p}_{a}^{l_{\!k\!}}\!-\!\mathbf{p}_{b}^{l_{\!k\!}})\!\times\!(\mathbf{p}_{a}^{l_{\!k\!}}\!-\!\mathbf{p}_{c}^{l_{\!k\!}}))|}{|(\mathbf{p}_{a}^{l_{k}}-\mathbf{p}_{b}^{l_{k}})\times(\mathbf{p}_{a}^{l_{k}}-\mathbf{p}_{c}^{l_{k}})|}\! & \text{if}~\mathbf{p}_{i}^{l_{k\!+\!1}}\!\in\!\mathbb{F}_{p}\!
\end{cases},
\end{equation}
and we have
\begin{equation}
\hat{\mathbf{p}}_{i}^{l_{k}}=\mathbf{R}_{l}^{b^{T}}(\mathbf{R}_{b_{k+1}}^{b_{k}}(\mathbf{R}_{l}^{b}\mathbf{p}_{i}^{l_{k+1}}+\mathbf{p}_{l}^{b})+\mathbf{p}_{b_{k+1}}^{b_{k}}-\mathbf{p}_{l}^{b}),
\end{equation}
where $\hat{\mathbf{p}}_{i}^{l_{k}}$ is the transformed point of $\mathbf{p}_{i}^{l_{k+1}}$ from $\mathcal{F}_{l_{k+1}}$ to $\mathcal{F}_{l_k}$. $\mathbf{R}_{l}^{b}$ and $\mathbf{p}_{l}^{b}$ together denote the extrinsic parameters between the lidar and IMU (calculated in off-line calibration). 

A physical explanation of Equation (\ref{Edge}) is provided in the followings. For an edge point, it describes the distance between $\hat{\mathbf{p}}_{i}^{l_{k}}$ and its corresponding edge $\overline{\mathbf{p}_{a}^{l_{k}}\mathbf{p}_{b}^{l_{k}}}$. For a planar point, it describes the distance between $\hat{\mathbf{p}}_{i}^{l_{k}}$ and its corresponding plane which is formed by three points, $\mathbf{p}_{a}^{l_{k}}$, $\mathbf{p}_{b}^{l_{k}}$, and $\mathbf{p}_{c}^{l_{k}}$. Details of how to choose $\mathbf{p}_{a}^{l_{k}}$, $\mathbf{p}_{b}^{l_{k}}$, and $\mathbf{p}_{c}^{l_{k}}$ can be found in \cite{zhang2014loam}.

We solve Equation (\ref{Cost Function}) using following iterated update equations:
\begin{align}
\mathbf{K}_{k,j}&=\mathbf{P}_{k}\mathbf{H}_{k,j}^{T}(\mathbf{H}_{k,j}\mathbf{P}_{k}\mathbf{H}_{k,j}^{T}+\mathbf{J}_{k,j}\mathbf{M}_{k}\mathbf{J}_{k,j}^{T})^{-1},\\
\Delta\mathbf{x}_{j}&=\mathbf{K}_{k,j}(\mathbf{H}_{k,j}\delta\mathbf{x}_{j}-f(^{-}\mathbf{x}_{b_{k+1}}^{b_{k}}\boxplus\delta\mathbf{x}_{j})),\\
\delta\mathbf{x}_{j+1}&=\delta\mathbf{x}_{j}+\Delta\mathbf{x}_{j},
\end{align}
where $\Delta\mathbf{x}_{j}$ denotes the correction vector at $j$th iteration. $\mathbf{H}_{k,j}$ is the jacobian of $f(^{-}\mathbf{x}_{b_{k+1}}^{b_{k}}\boxplus\delta\mathbf{x}_{j})$ w.r.t. $\delta\mathbf{x}_j$. Note that, in every iteration, we will find new matched edges and planes to further minimize the error metric, followed by computing new $\mathbf{H}_{k,j}$, $\mathbf{J}_{k,j}$, and $\mathbf{K}_{k,j}$. When $f(\mathbf{x}_{b_{k+1}}^{b_{k}})$ is below a certain threshold, say at the $n$ th iteration, we update $\mathbf{P}_{k}$ by
\begin{equation}
\mathbf{P}_{\!k\!+\!1\!}\!=\!(\!\mathbf{I}\!-\!\mathbf{K}_{k,n}\mathbf{H}_{k,n}\!)\mathbf{P}_{k}(\!\mathbf{I}\!-\!\mathbf{K}_{k,n}\mathbf{H}_{k,n}\!)^{T}\!+\!\mathbf{K}_{k,n}\mathbf{M}_{k}\mathbf{K}_{k,n}^{T}.
\end{equation}

Using Equation (\ref{Box plus}), we are able to obtain the final $\mathbf{x}_{b_{k+1}}^{b_{k}}$. The raw distorted features can now be undistorted using the estimated relative transformation.

Finally, we initialize the next state, $\mathbf{x}_{b_{k+2}}^{b_{k+1}}$, with
\begin{equation}
[\mathbf{0}_{3},\mathbf{v}_{b_{k+1}}^{b_{k+1}},\mathbf{q}_{0},\mathbf{b}_{a},\mathbf{b}_{g},\mathbf{g}^{b_{k+1}}],
\end{equation}
where $\mathbf{q}_{0}$ denotes identity quaternion. $\mathbf{v}_{b_{k+1}}^{b_{k+1}}$ and $\mathbf{g}^{b_{k+1}}$ can be computed by $\mathbf{v}_{b_{k+1}}^{b_{k+1}}=\mathbf{R}_{b_{k}}^{b_{k+1}}\mathbf{v}_{b_{k+1}}^{b_{k}}$ and $\mathbf{g}^{b_{k+1}}=\mathbf{R}_{b_{k}}^{b_{k+1}}\mathbf{g}^{b_{k}}$, respectively. Note that, the covariances regarding the velocity, biases, and local gravity remain in the covariance matrix, while the covariance corresponding to the relative pose is set to zero, i.e., no uncertainty for
the robocentric frame of reference itself. 

\subsubsection{State Composition}
In the robocentric formulation, every time when the update is finished, we need to update the global pose, $\mathbf{x}_{w}^{b_k}$, through a composition step as
\begin{equation}
\mathbf{x}_{w}^{b_{k+1}}=\left[\begin{array}{c}
\mathbf{p}_{w}^{b_{k+1}}\\
\mathbf{q}_{w}^{b_{k+1}}
\end{array}\right]=\left[\begin{array}{c}
\mathbf{R}_{b_{k}}^{b_{k+1}}(\mathbf{p}_{w}^{b_{k}}-\mathbf{p}_{b_{k+1}}^{b_{k}})\\
\mathbf{q}_{b_{k}}^{b_{k+1}}\otimes\mathbf{q}_{w}^{b_{k}}
\end{array}\right].
\end{equation}

\subsubsection{Initialization}
As described in Sect. \ref{Lidar-Inertial Odometry with IESKF}, the robocentric formulation can facilitate the initialization of the filter state. Regarding the initial parameter settings, in our implementation, (i) the initial acceleration bias and lidar-IMU extrinsic parameters are obtained via off-line calibration, while the initial gyroscope bias is the average of the corresponding stationary measurements, (ii) the initial roll and pitch are obtained from the unbiased acceleration measurements before moving, and (iii) the initial local gravity is acquired via transforming the gravity vector represented in navigation frame to current local frame using initial roll and pitch from (ii).

\begin{figure*}[t]
	\centering
	\subfigure[Port area]{\includegraphics[height=0.85in]{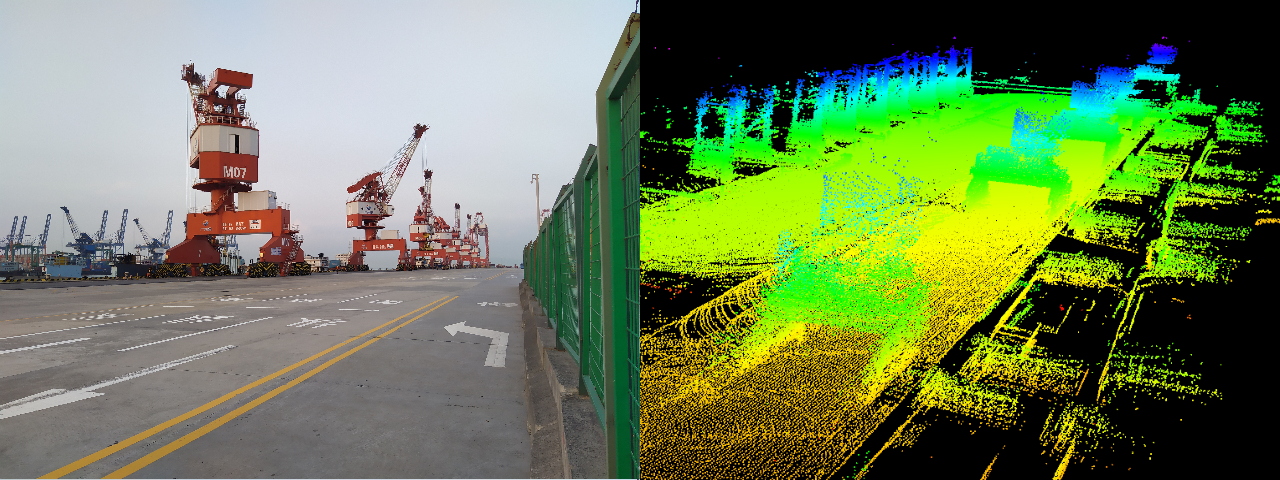}\label{fig_port_merge}}
	\subfigure[Industrial park area]{\includegraphics[height=0.85in]{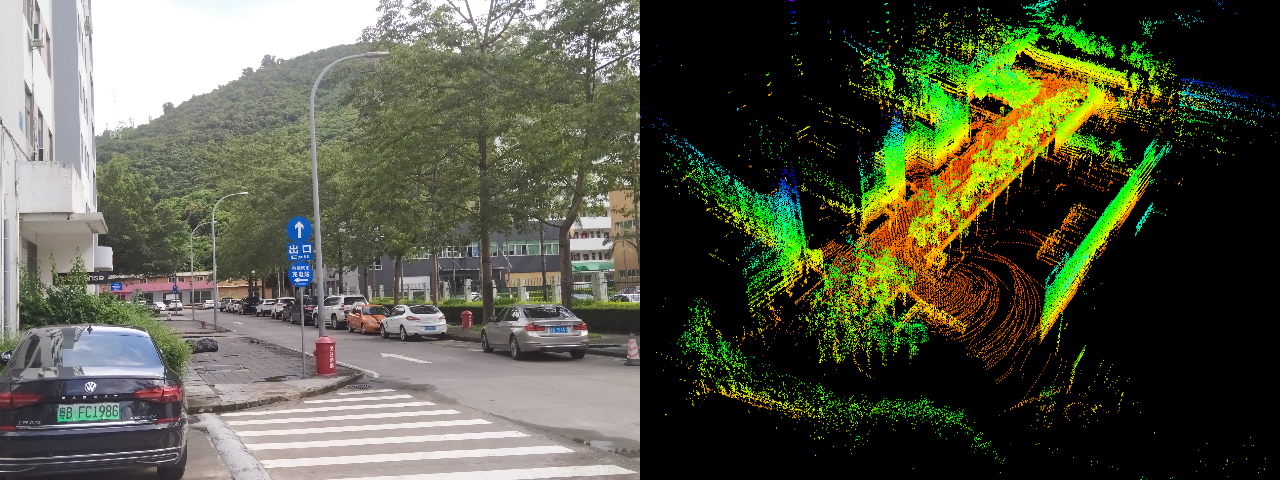}\label{fig_city_merge}}
	\subfigure[Forest area]{\includegraphics[height=0.85in]{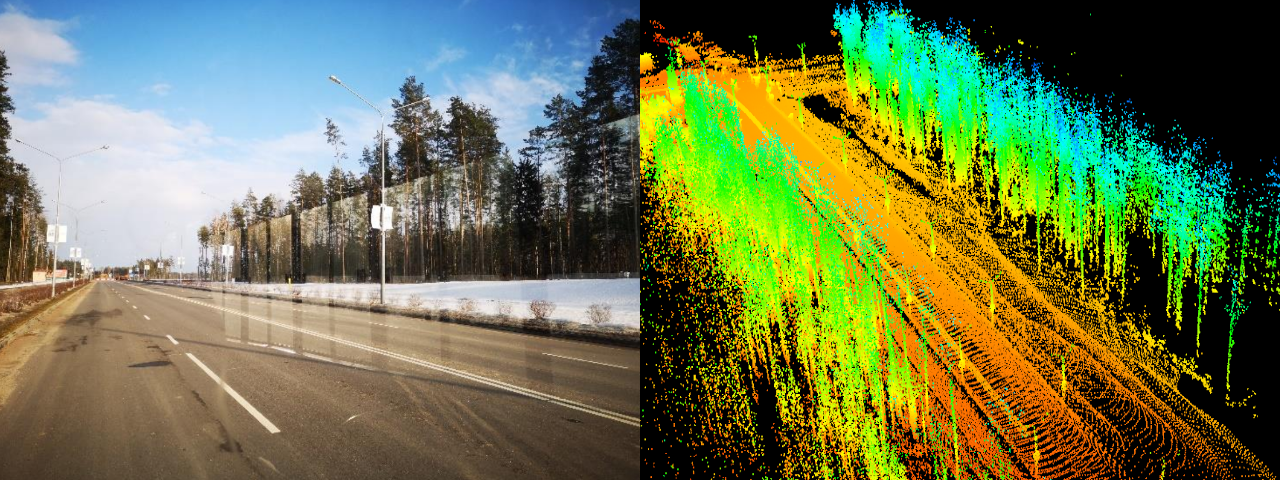}\label{fig_forest_merge}}
	\caption{Photos and corresponding maps (produced by LINS) of (a) a wide and open port area, (c) an industrial park with numerous buildings, trees, and cars, and (d) a clean road through a forest area. LINS performs well in all tested environments.}	
	\label{fig_many_scenarios}
\end{figure*}

\begin{table*}[t]
	\caption{Reative errors for motion estimation drift}
	\label{tab_1}
	\centering
	\begin{tabular}{@{}cccccccccccc@{}}
		\toprule
		&                                & \multicolumn{2}{c}{Num. of Features} & \multicolumn{4}{c}{Drifts of Map-Refined Odometry (\%)}                                                                                                                                 & \multicolumn{4}{c}{Drifts of Pure Odometry (\%)}             \\ \midrule
		\multicolumn{1}{c|}{Scenario}     & \multicolumn{1}{c|}{Dist. (m)} & Edge  & \multicolumn{1}{c|}{Planar} & LOAM \cite{zhang2014loam} & LeGO \cite{shan2018lego} & LIOM \cite{ye2019tightly} & \multicolumn{1}{c|}{LINS}          & LOAM  & LeGO  & LIOM          & LINS          \\ \midrule
		\multicolumn{1}{c|}{City}       & \multicolumn{1}{c|}{1100}      & 85    & \multicolumn{1}{c|}{2552}   & 72.91                                      & 10.17                                     & \textbf{1.76}                              & \multicolumn{1}{c|}{1.79}          & 76.84 & 30.13 & 4.44          & \textbf{4.42} \\ \midrule
		\multicolumn{1}{c|}{Port}        & \multicolumn{1}{c|}{1264}      & 103   & \multicolumn{1}{c|}{2487}   & 2.16                                       & 3.35                                      & \textbf{1.40}                                       & \multicolumn{1}{c|}{1.56} & 4.64  & 8.70  & \textbf{1.72} & 2.75          \\ \midrule
		\multicolumn{1}{c|}{Park}        & \multicolumn{1}{c|}{117}       & 420   & \multicolumn{1}{c|}{3598}   & 19.35                                      & 1.97                                      & 2.61                                       & \multicolumn{1}{c|}{\textbf{1.32}} & 26.50 & 26.08 & 13.60         & \textbf{7.69} \\ \midrule
		\multicolumn{1}{c|}{Forest}      & \multicolumn{1}{c|}{371}       & 99    & \multicolumn{1}{c|}{2633}   & 5.59                                       & 3.66                                      & 9.58                                       & \multicolumn{1}{c|}{\textbf{3.31}} & 10.60 & 18.93 & 12.96         & \textbf{7.27} \\ \midrule
		\multicolumn{1}{c|}{Parking Lot} & \multicolumn{1}{c|}{144}       & 512   & \multicolumn{1}{c|}{5555}   & 1.21                                       & 1.12                                      & \textbf{1.05}                              & \multicolumn{1}{c|}{1.08}          & 5.38  & 6.62  & 2.17          & \textbf{1.72} \\ \bottomrule
	\end{tabular}
\end{table*}

\section{Experiments}\label{Experiments}
We now evaluate the performance of LINS in different scenarios and compare it with LeGO \cite{shan2018lego}, LOAM \cite{zhang2014loam}, and LIOM \cite{ye2019tightly}, on a laptop computer with 2.4GHz quad cores and 8Gib memory. All methods are implemented in C++ and executed using the robot operating system (ROS) \cite{quigley2009ros} in Ubuntu Linux. In the following experiments, the mapping module of LINS is implemented by the mapping algorithm proposed in LeGO \cite{shan2018lego}. Most of the previous works only analyzed the performance of the final trajectory, i.e. the odometry already refined by the map. However, we found that the initial odometry, i.e. the odometry purely produced by the odometry module, has great impact to the overall performance. Therefore, we took both of them into account. To distinguish these two odometries, we termed the odometry refined by the map as map-refined odometry (MRO), and the initial odometry as pure odometry (PO).

\begin{figure}[b]
	\centering
	\subfigure[]{\includegraphics[height=1.0in]{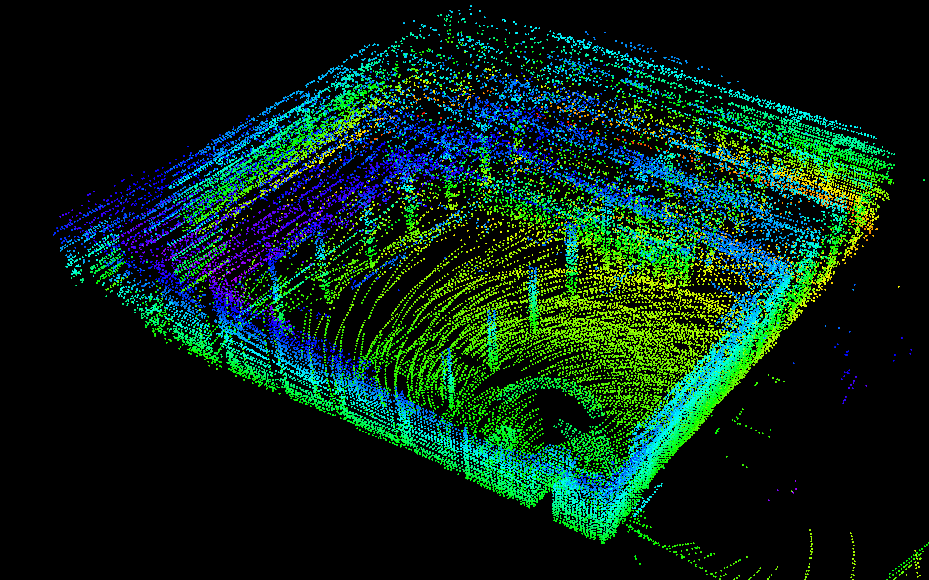}\label{fig_parking_lot}}
	\subfigure[]{\includegraphics[height=1.0in]{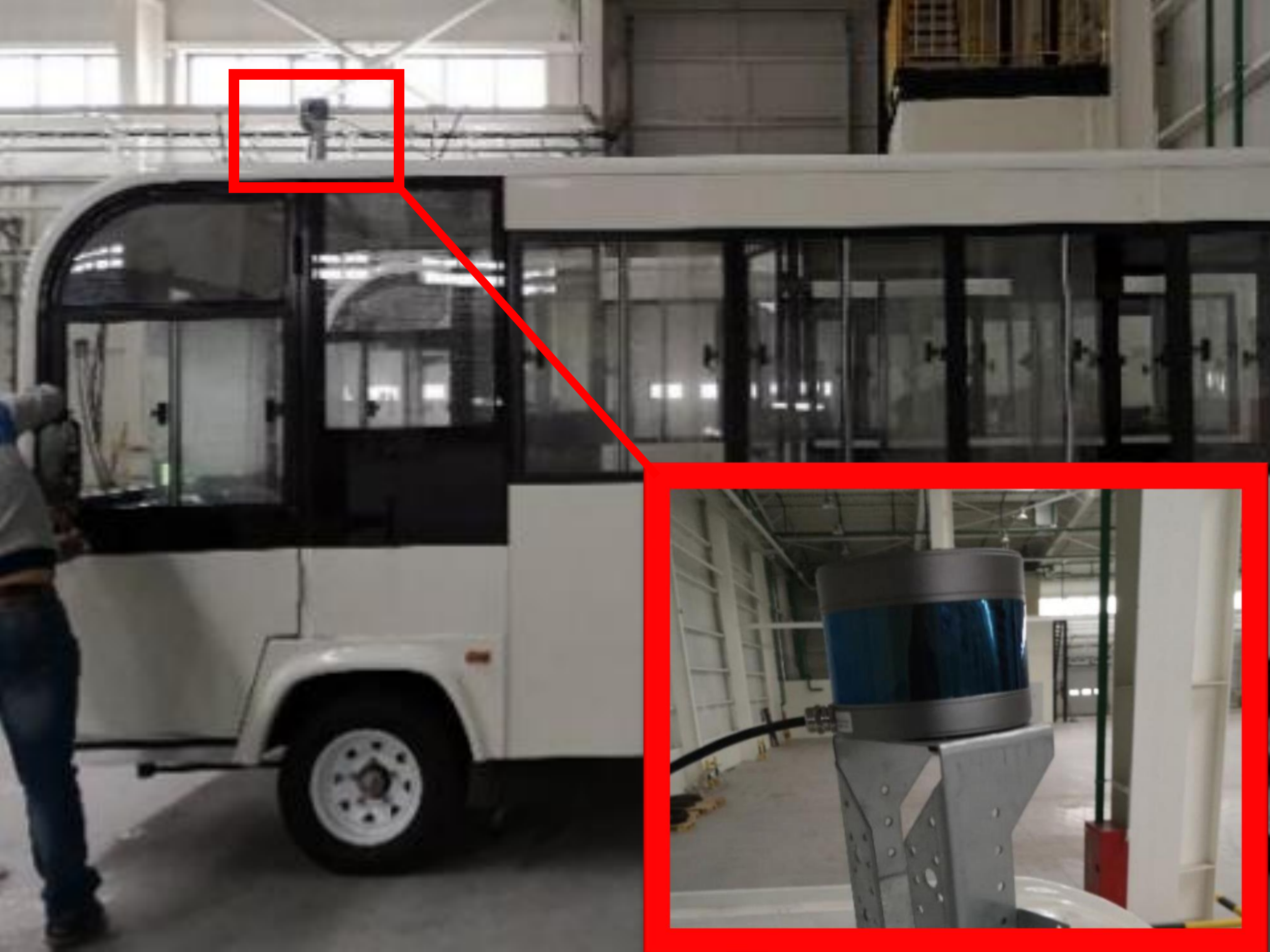}\label{fig_bus}}	
	\caption{The sensor configuration for indoor tests. (a) Map of the parking lot built by LINS. (b) Lidar installation. An IMU is stuck to the bus.}	
\end{figure}

\begin{figure}[b]
	\centering	
	\subfigure[LINS]{\includegraphics[height=0.825in]{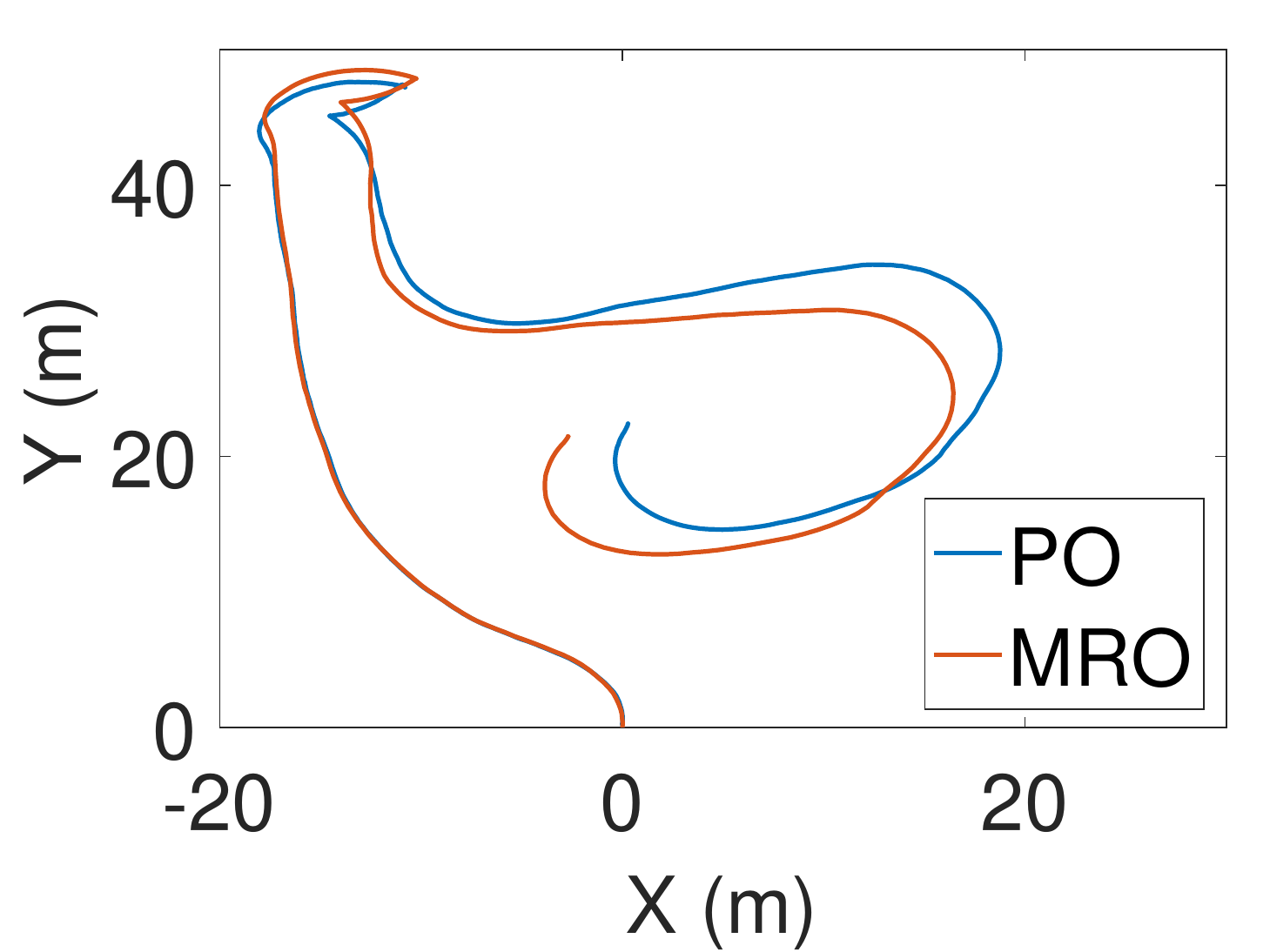}\label{fig_exp_indoor_lins_1}}
	\subfigure[LeGO]{\includegraphics[height=0.825in]{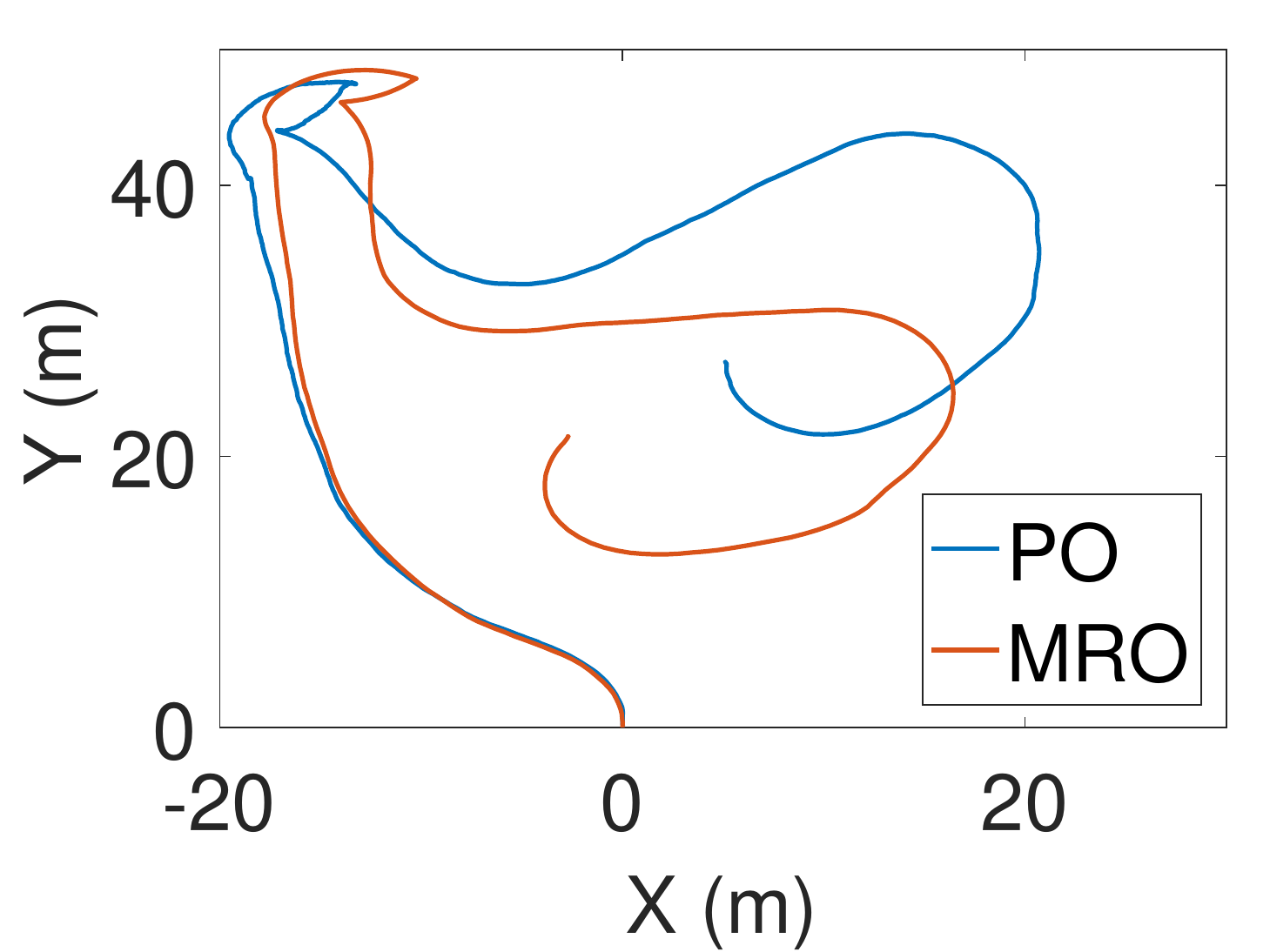}\label{fig_exp_indoor_lego_1}}	
	\subfigure[LOAM]{\includegraphics[height=0.825in]{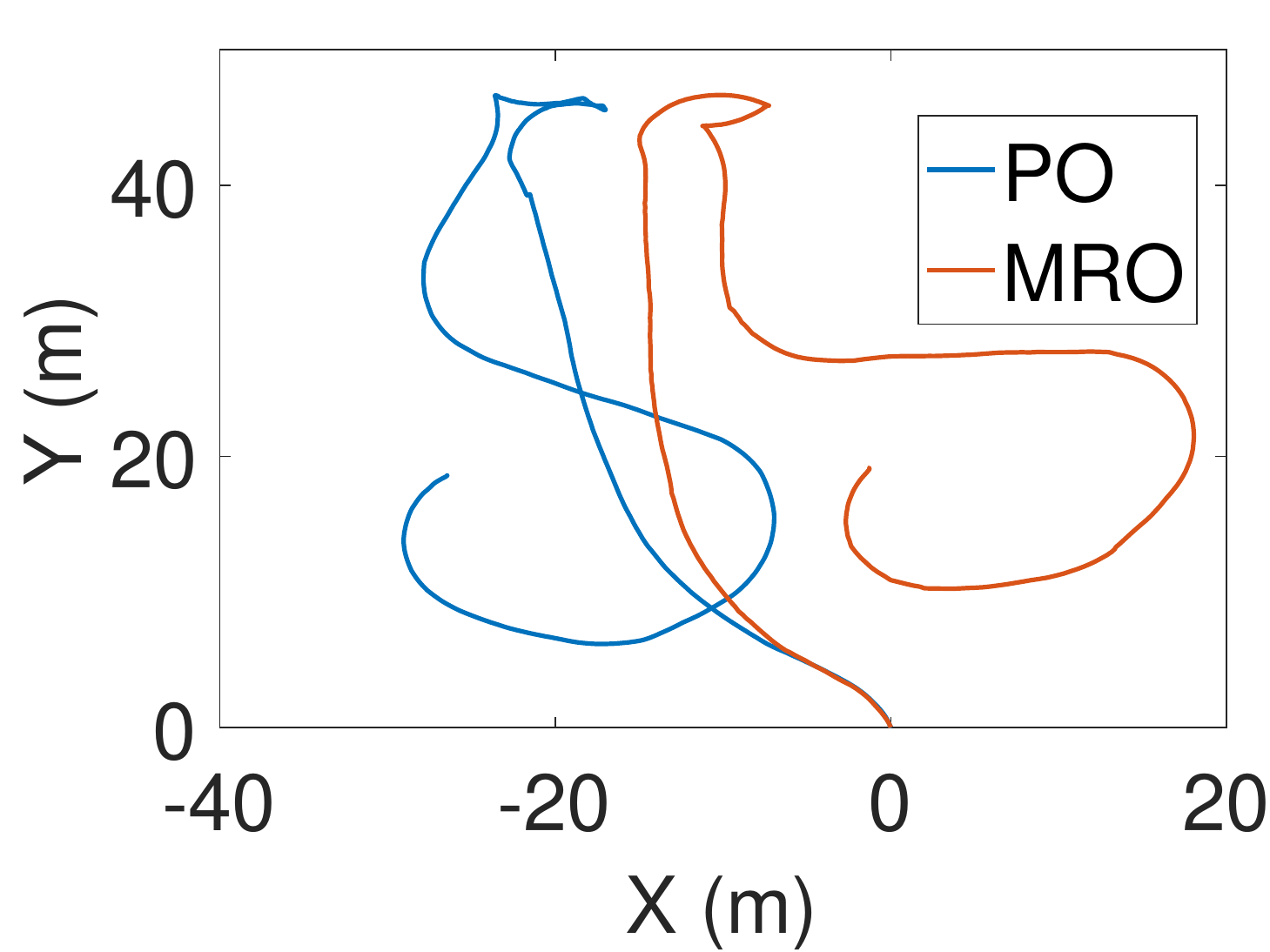}\label{fig_exp_indoor_loam_1}}	
	\caption{Resulting trajectories in the indoor experiment of (a) LINS, (b) LeGO, and (c) LOAM. We observe that MRO trajectories from all methods look very similar, but their PO trajectories are totally different. The PO trajectory from LINS aligns much better with its MRO trajectory than other methods.}
\end{figure}

\subsection{Indoor Experiment}
In the indoor test, a parking lot was chosen as the experiment area as shown in Fig. \ref{fig_parking_lot}. We installed our sensor suite on a bus as shown in Fig. \ref{fig_bus}, where a RS-LiDAR-16 was mounted on the top and an IMU was placed inside the bus. Fig. \ref{fig_exp_indoor_lins_1}, \ref{fig_exp_indoor_lego_1}, and \ref{fig_exp_indoor_loam_1} provide the results from LINS, LeGO, and LOAM, respectively. Although we do not have ground truth, we can still visually inspect that LINS-PO's trajectory can be precisely aligned with the MRO trajectory (generally, MRO is almost drift-free indoors and more accurate than PO), while the LeGO-PO and LIOM-PO both have noticeable drifts in the yaw angle.

\subsection{Large-scale Outdoor Environment}
To verify generalizability and stability, experiments were carried out in four outdoor application scenarios: city, port, industrial park, and forest. Fig. \ref{fig_many_scenarios} showcases some photos of the environments and corresponding maps generated by LINS. We measured the gap between the ground truth produced by a GPS receiver and the estimated position provided by each method, which indicates the amount of drift, and then compare it to the distance traveled to yield a relative drift. The experimental results are listed in Table \ref{tab_1}.

In summary, LINS performed well in all tested scenarios. The detailed analysis for specific environments is conducted below.

\subsubsection{Port Experiment}
We evaluated LINS in a port in Guangdong. The sensor suite consisted of a Velodyne VLP-16 lidar and an Xsens MTi-G-710 IMU fixed on the top of a car. The ground-truth trajectory was offered by a GPS module. We started recording data from a path surrounded by containers. The car headed to a dock and then returned to the original spot after traveling a distance of 1264 meters. It is worth to mention that the containers would go in and out incessantly changing the global map, which may undermine the performance of MRO.

\begin{figure}[t]
	\centering
	\subfigure[Trajectories from LeGO]{\includegraphics[height=1.20in]{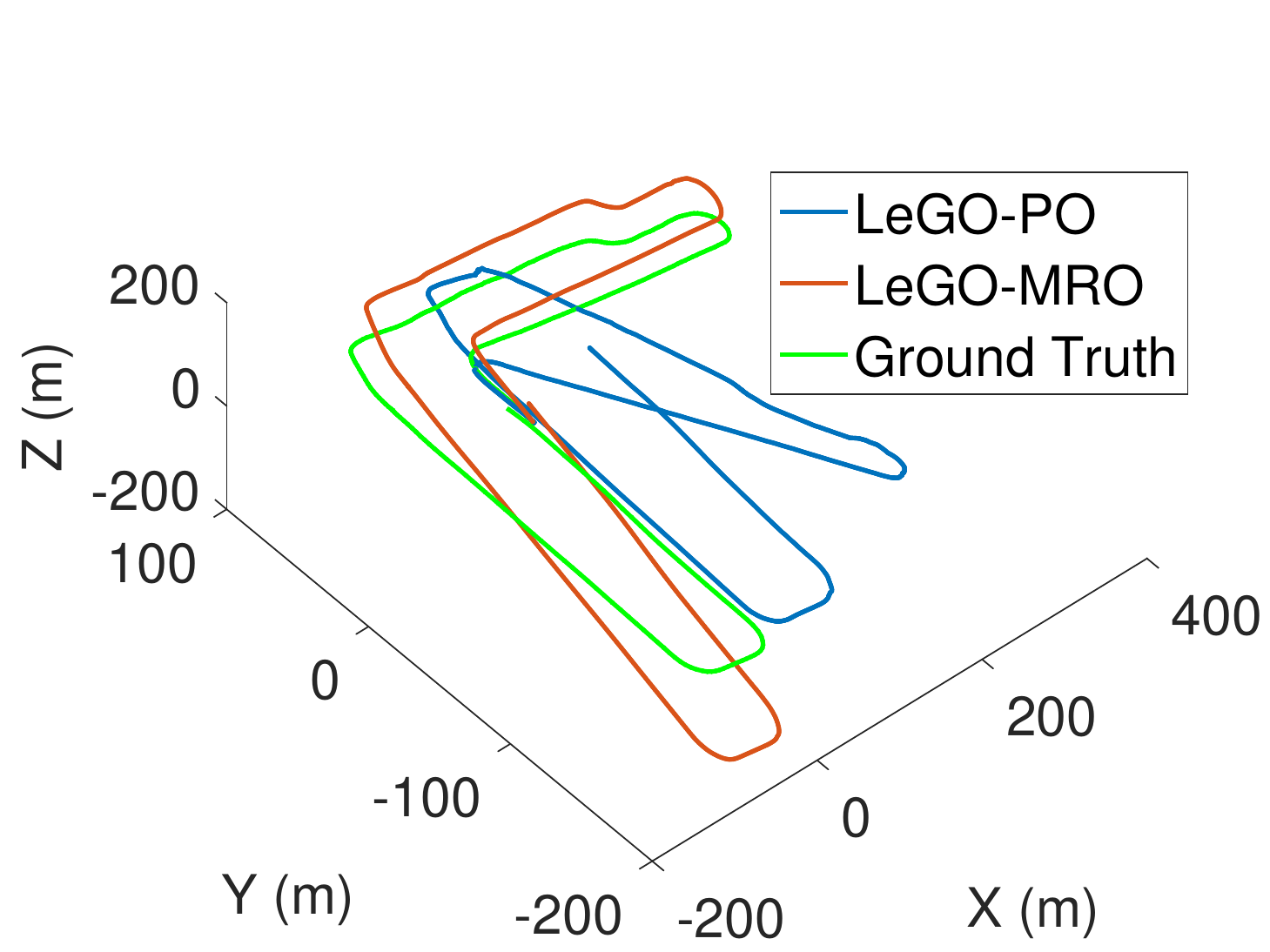}\label{fig_exp_port_lego_1}}
	\subfigure[Trajectories from LINS]{\includegraphics[height=1.20in]{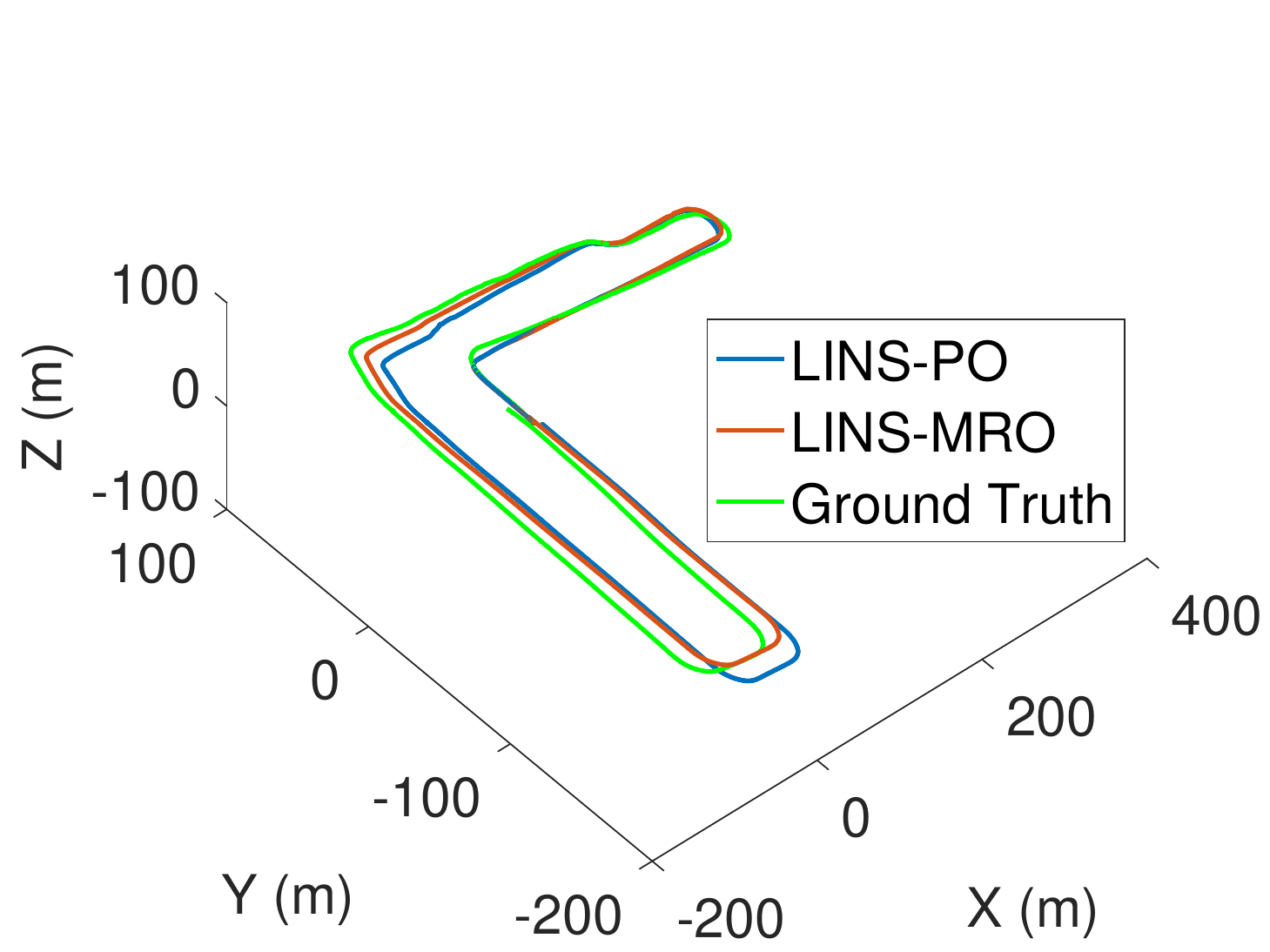}\label{fig_exp_port_lins_1}}	
	\subfigure[Map built by LeGO]{\includegraphics[height=1.1in]{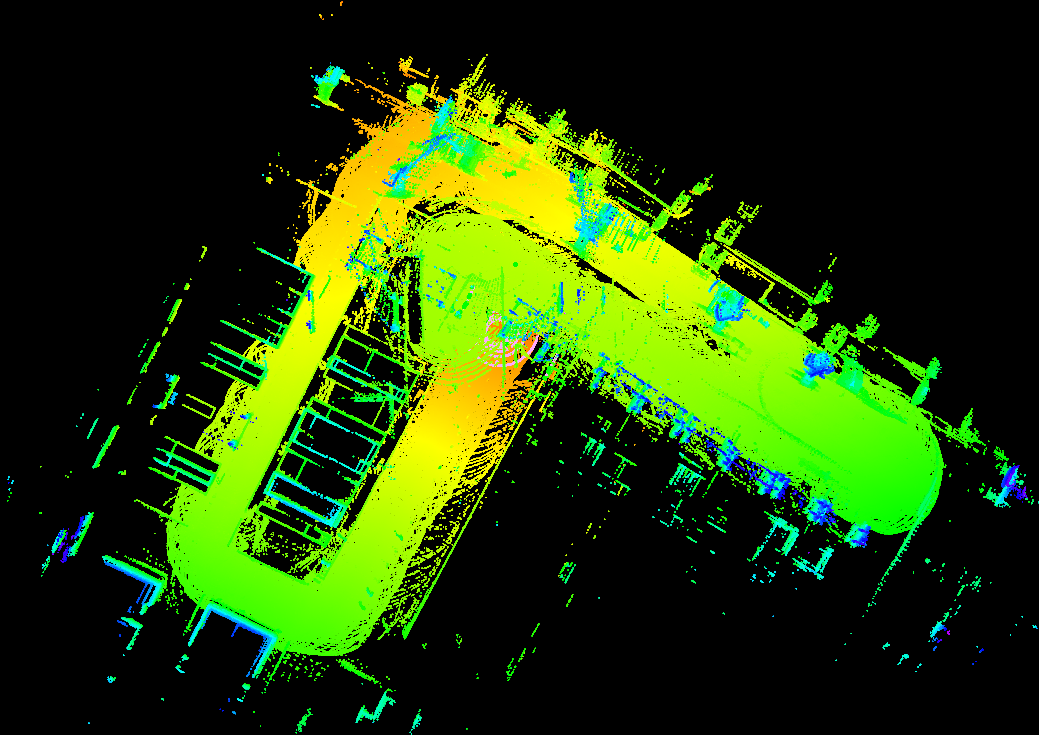}\label{fig_lego_pcl_exp2}}	
	\subfigure[Map built by LINS]{\includegraphics[height=1.1in]{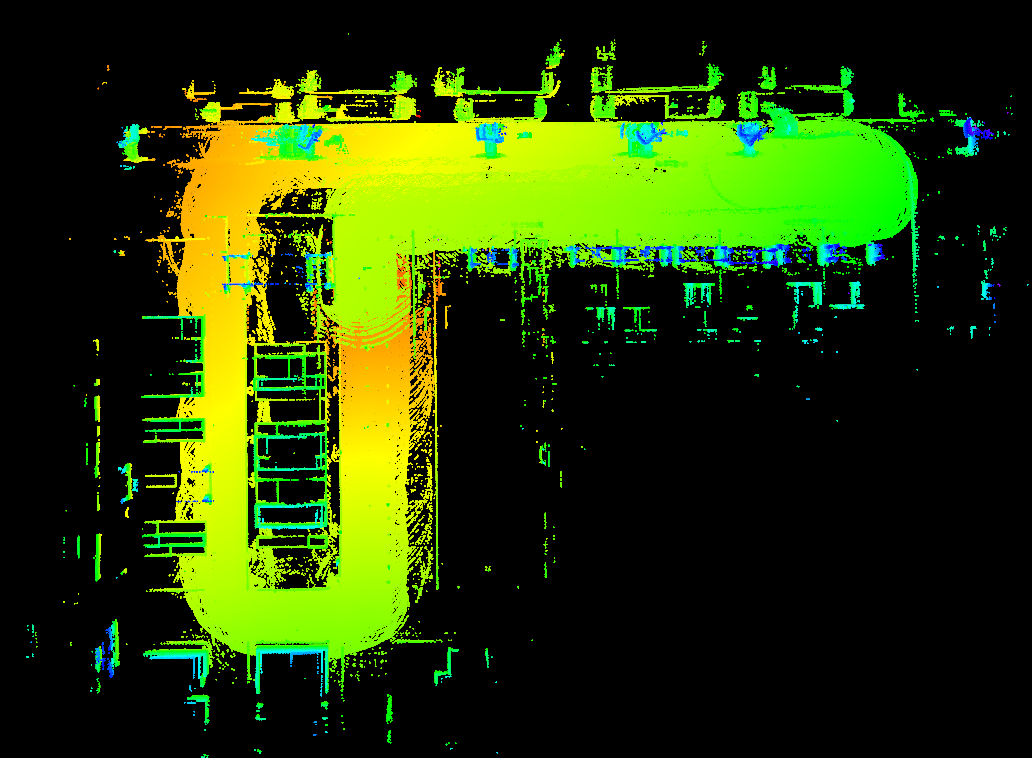}\label{fig_lins_pcl_exp2}}		
	\caption{Estimated trajectories and maps from LeGO and LINS. Note that the PO trajectories are drawn in blue lines, the MRO trajectories are drawn by red lines, and the GPS ground truthes are drawn in green lines. We see that the trajectories from LINS is close to the ground truthes, and the resulting map has higher fidelity than that of LeGO.}	
\end{figure}

According to Table \ref{tab_1}, we find that LINS and LIOM present the lowest drifts. The relative drift of LINS-MRO is 1.56\%, slightly higher than that of LIOM which is 1.40\%, while the relative drift of LINS-PO is just 2.75\%. The results indicate that combining IMU and lidar can effectively improve accuracy. Even though the relative drifts of LOAM and LeGO seems to be small, they may suffer from huge orientation errors. Fig. \ref{fig_exp_port_lego_1} and \ref{fig_exp_port_lins_1} provide detailed trajectories and maps from LeGO and LINS. Compared to the ground truths (green lines), we find trajectories from LeGO (including MRO and PO) turned to an erroneous direction around the first turn. We can also visually inspect the deformation of the map built by LeGO in Fig. \ref{fig_lego_pcl_exp2}. In contrast, LINS exhibited good alignment with the ground-truth trajectory and the resulting map presented high fidelity to the real-world environment. Even in the first turn where features were insufficient (only about 30 edge features available per scan), LINS performed very well, which shows that our algorithm is more robust to feature-less scenes.

\begin{table}[htbp]
	\caption{Runtime of the LIO module per scan}
	\label{tab_2}
	\centering
	\begin{tabular}{@{}cccccc@{}}
		\toprule
		Method & City & Port  & Park  & Forest & Parking Lot \\ \midrule
		LIOM   & 143 ms & 185 ms & 201 ms & 173  ms & 223 ms      \\
		LINS   & \textbf{18 ms} & \textbf{19 ms} & \textbf{21 ms} & \textbf{20 ms}  & \textbf{25 ms}       \\ \bottomrule
	\end{tabular}
\end{table}

\begin{figure}[t]
	\centering
	\subfigure[LINS's Trajectory overlaid with Google Map for visual comparison]{\includegraphics[width=0.35\textwidth]{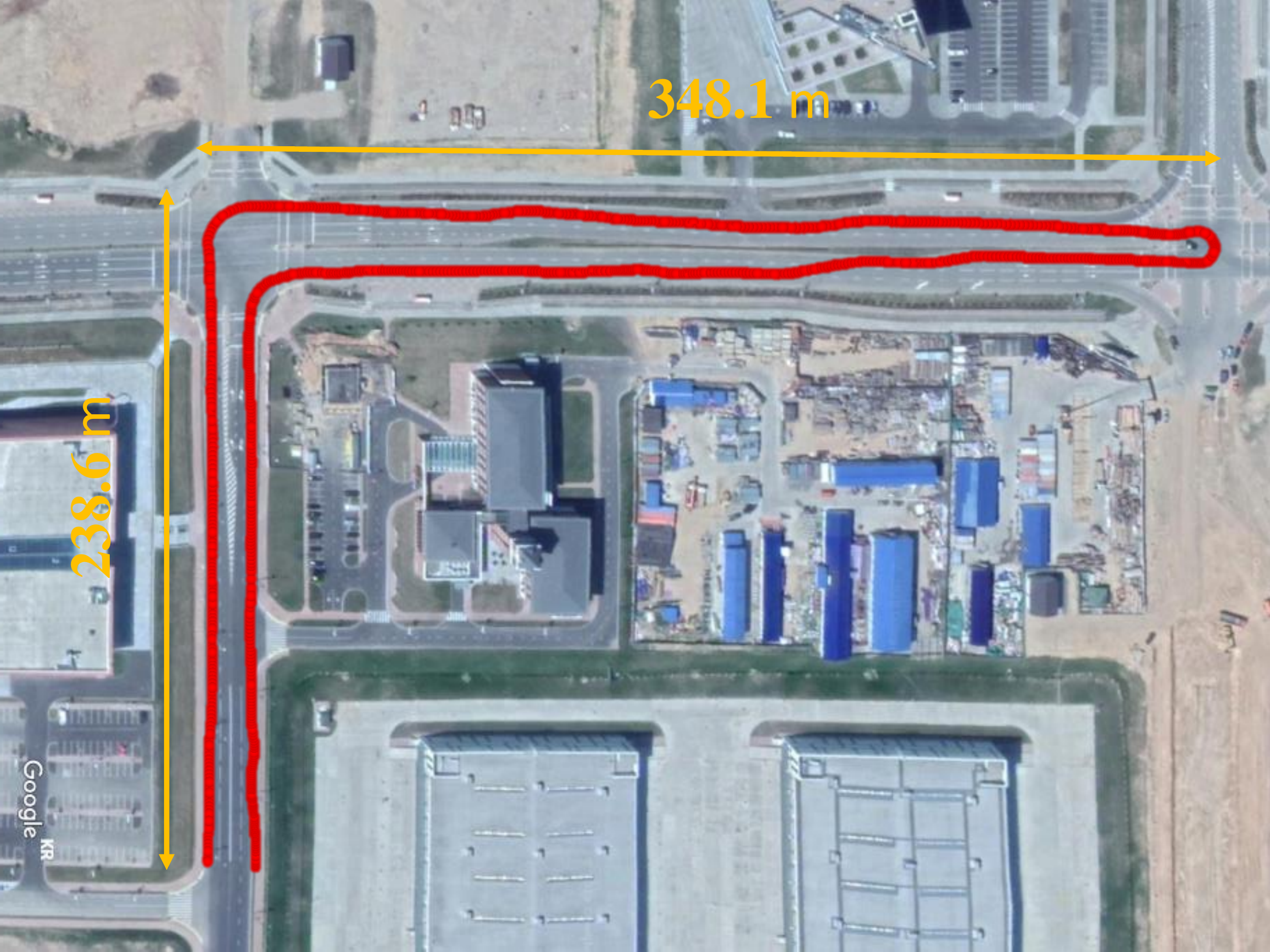}\label{fig_google_earth_urban}}	
	\subfigure[LeGO]{\includegraphics[width=0.145\textwidth]{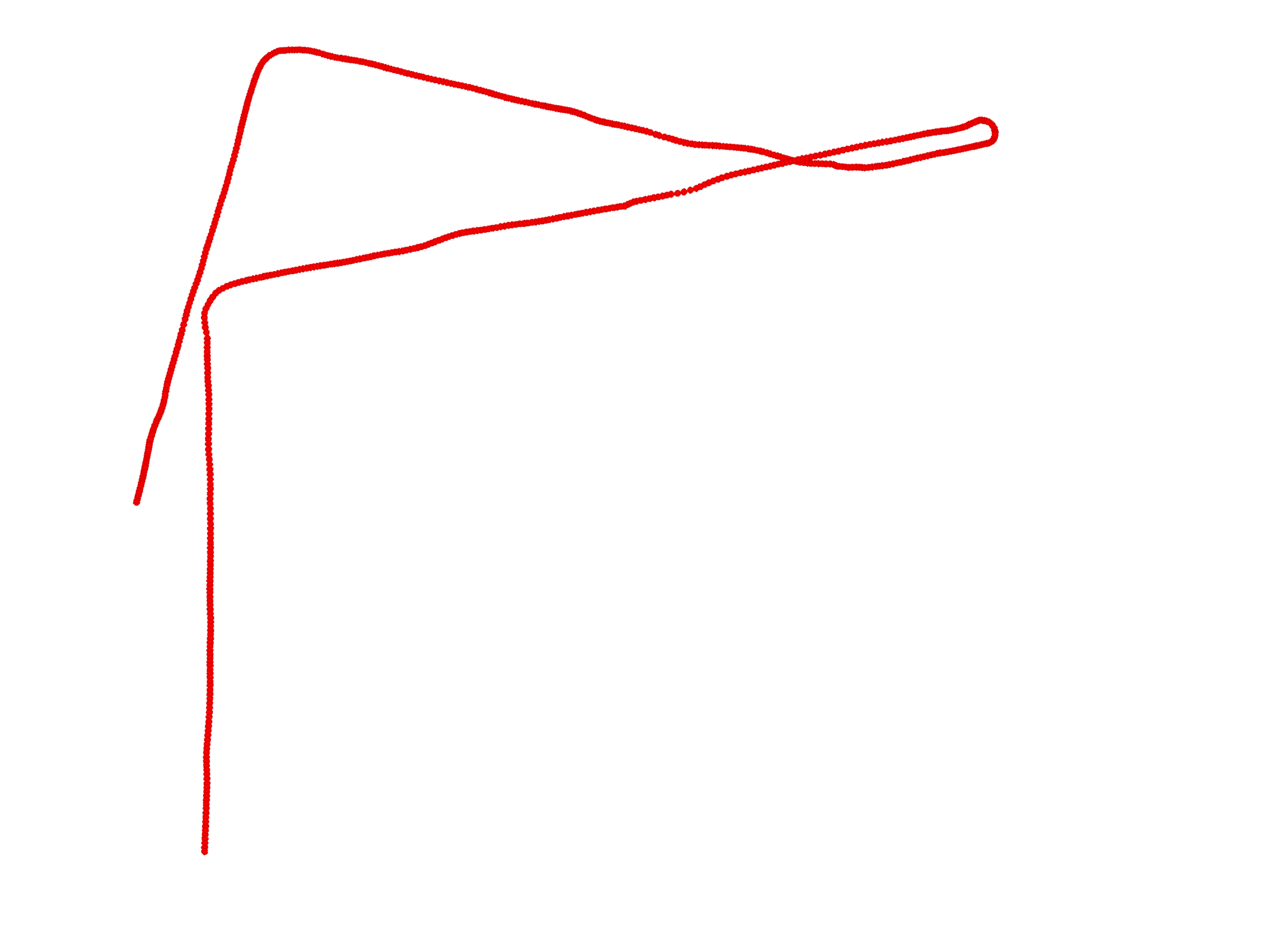}\label{fig_exp_urban_lego}}	
	\subfigure[LOAM]{\includegraphics[width=0.16\textwidth]{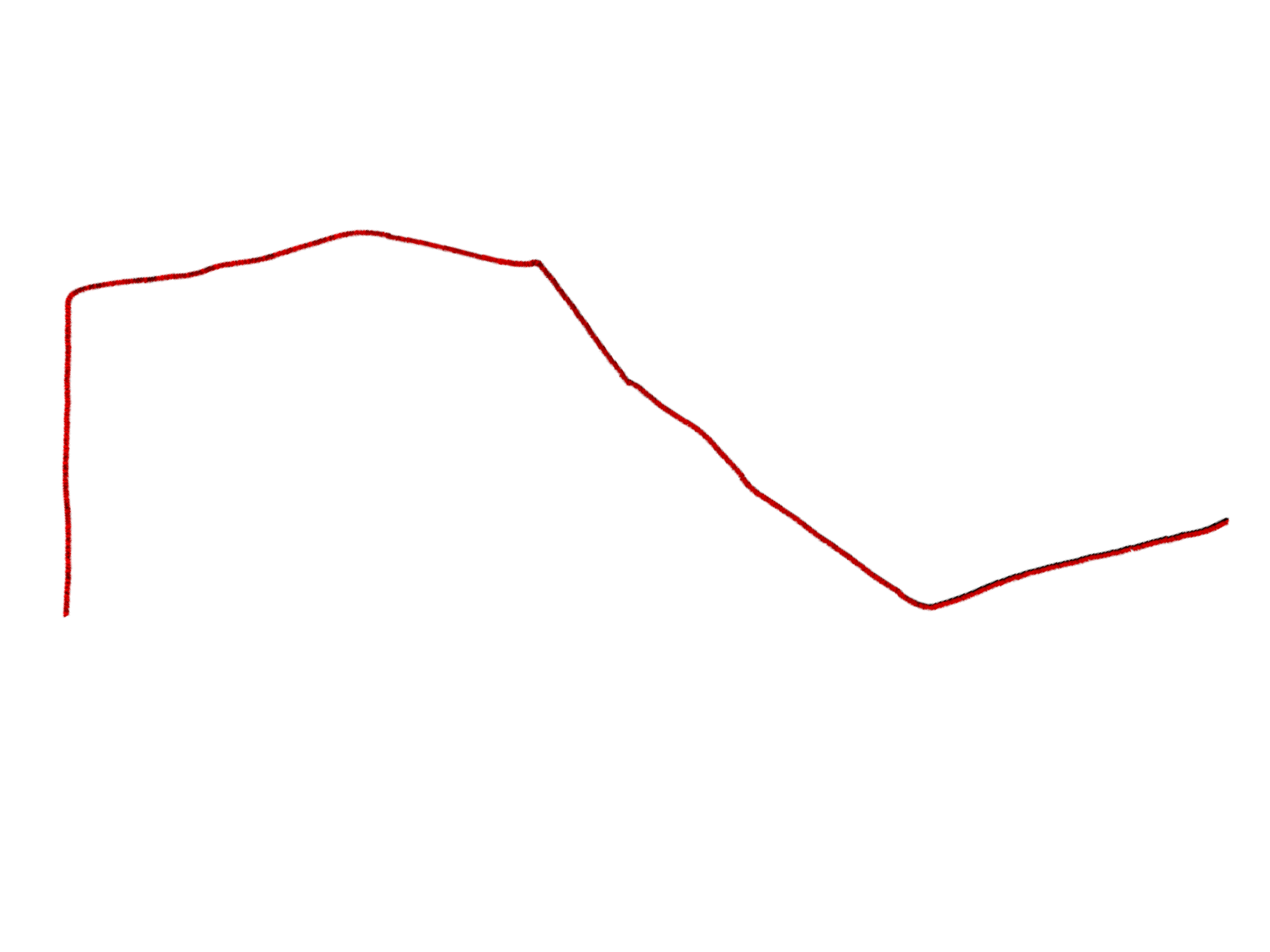}\label{fig_exp_urban_loam}}
	\subfigure[LIOM]{\includegraphics[width=0.145\textwidth]{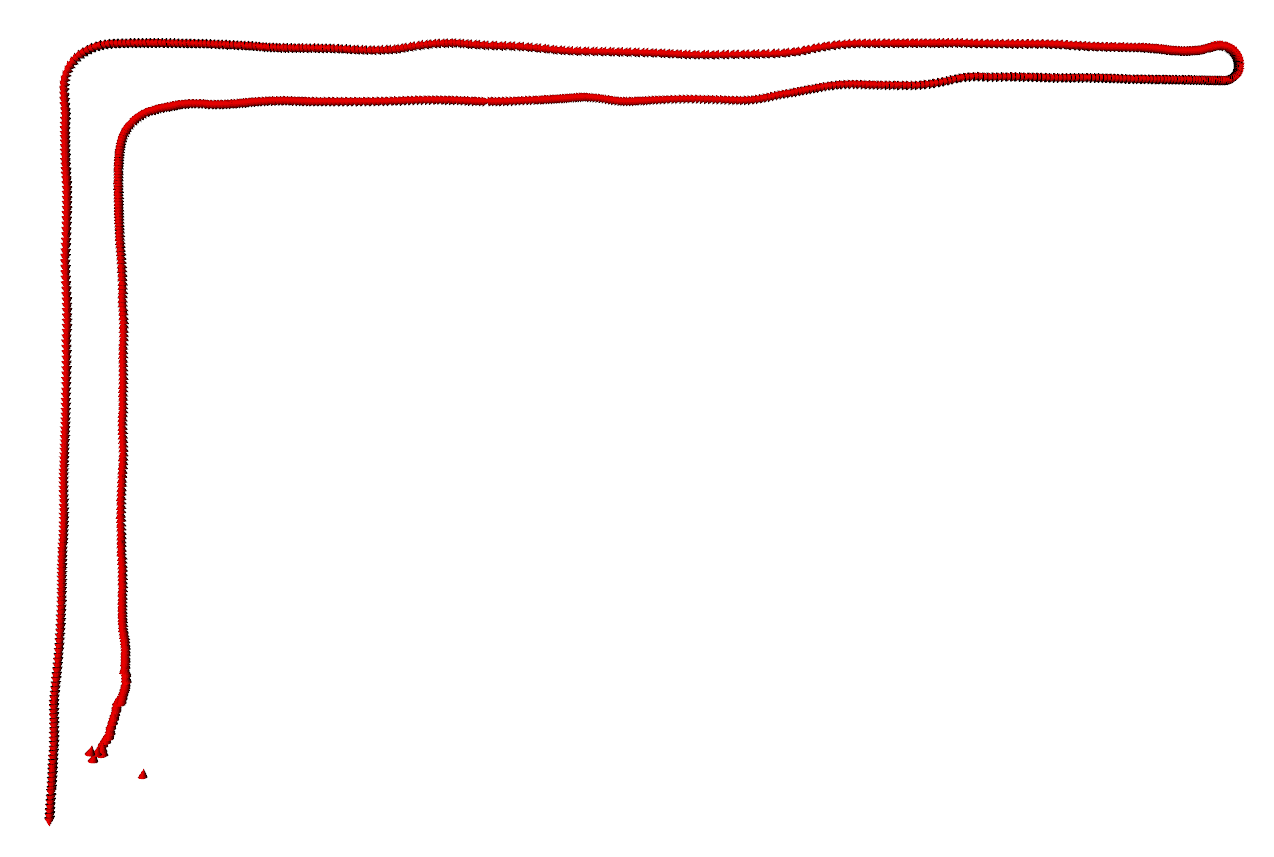}\label{fig_exp_urban_liom}}
	\caption{MRO trajectories generated by different methods in the urban experiment, which are drawn in red lines.}
\end{figure}

\begin{figure}[t]
	\centering	
	\subfigure[ATE of MRO]{\includegraphics[width=0.225\textwidth]{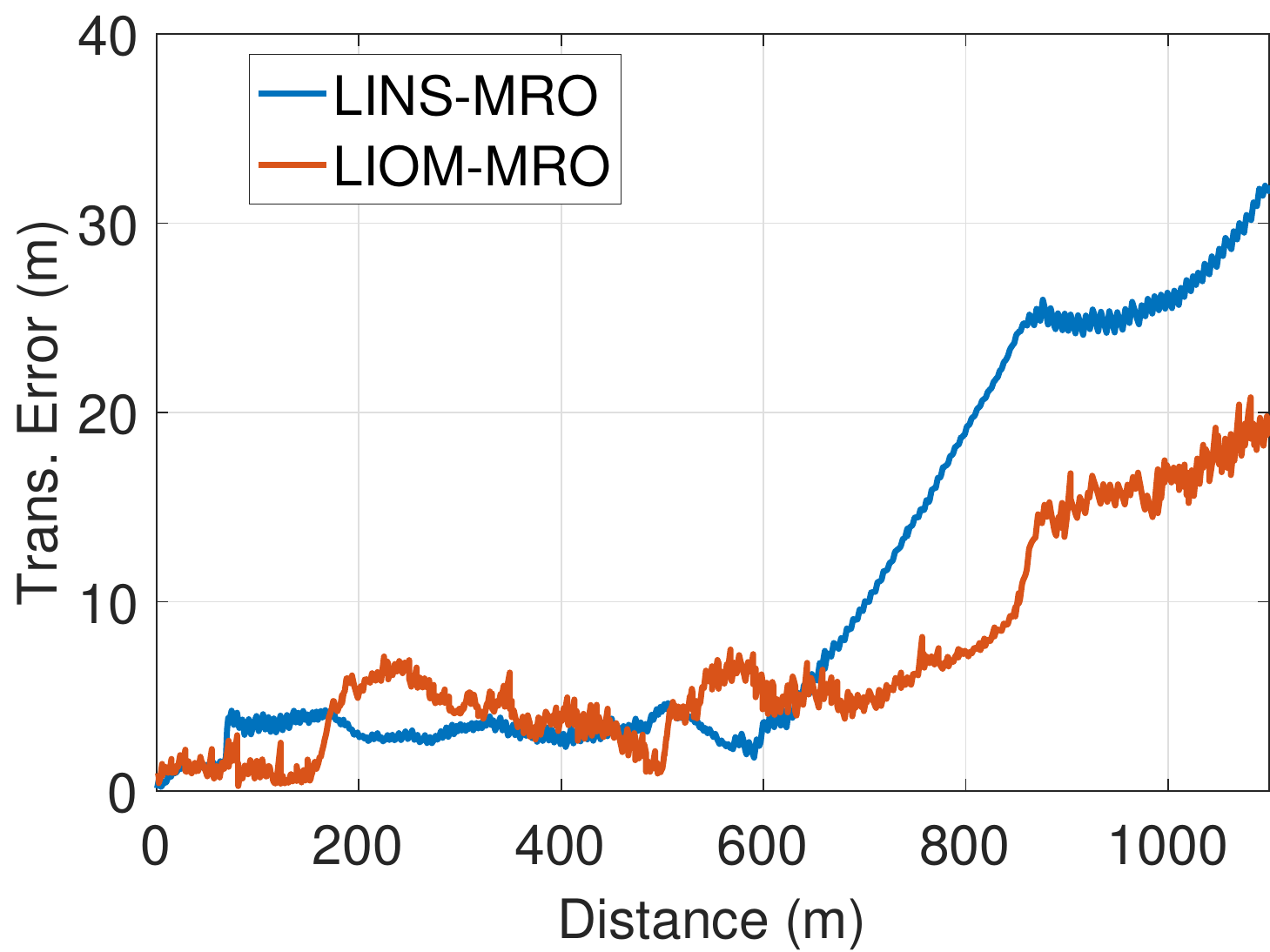}\label{fig_exp_urban_trans_error_1}}	
	\subfigure[ATE of PO]{\includegraphics[width=0.225\textwidth]{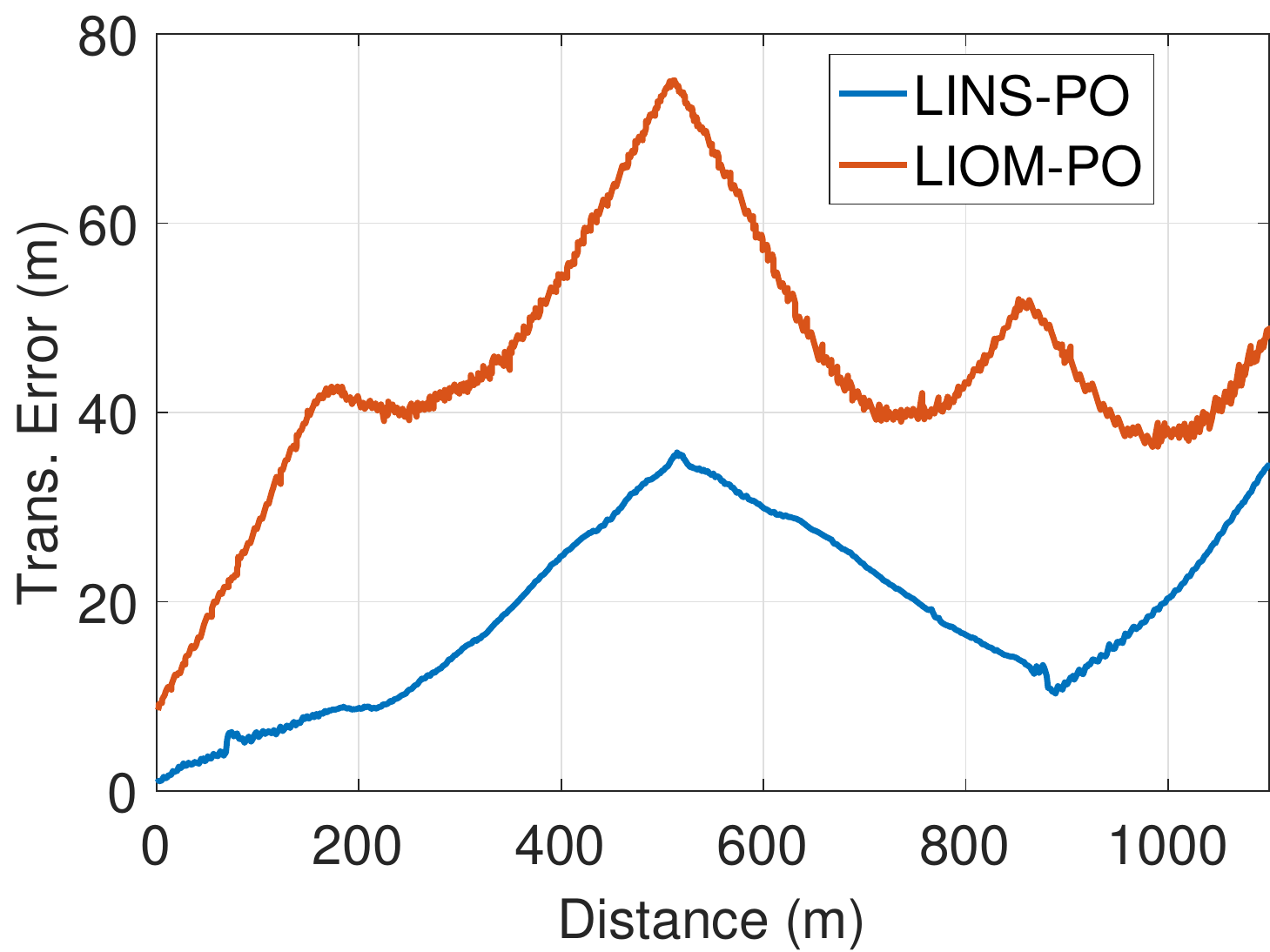}\label{fig_exp_urban_trans_error_2}}
	\caption{Comparison of ATE bewteen LINS and LIOM w.r.t MRO and PO.}	
\end{figure}

\subsubsection{Urban Experiment}
We carried out the urban experiment with the same sensor suite in Fig. \ref{fig_bus}. Positions produced by a GPS receiver were used as ground truthes. It is worth to mention that in this scene, the average edge feature number is only 56 per scan, which is the lowest among all tested scenarios. 

We first take a glance at the output of LeGO and LOAM, which can be seen in Fig. \ref{fig_exp_urban_lego} and \ref{fig_exp_urban_loam}, respectively. We observed that huge orientation errors occur in almost every turns. Fig. \ref{fig_google_earth_urban} showcases the result of LINS when running on the same dataset. The resulting trajectories show good superposition with the real-world road, which confirms that LINS can stably run even in the feature-less scenes. The final MRO and PO drifts of LINS are 1.79\% and 4.42\%, respectively, very close to that of LIOM which are 1.76\% and 4.44\%, respectively. Moreover, in comparison to the trajectory from LIOM as shown in Fig. \ref{fig_exp_urban_liom} and absolute trajectory error (ATE) as shown in Fig. \ref{fig_exp_urban_trans_error_1} and \ref{fig_exp_urban_trans_error_2}, we can see the performance of LINS is close to that of LIOM in terms of accuracy. We analyze that LIOM benefits from rotation-constrained refinement in the mapping step, which results in higher accuracy in the MRO results.

\subsection{Runtime Comparison}
Table \ref{tab_2} compares the mean runtimes of the lidar-inertial odometry module in LINS and LIOM. We see that LINS is much faster than LIOM---LINS requires less than 30 millisecond in processing one scan, while LIOM always requires more than 100 millisecond. In some extreme case such as the parking lot where features abound, LIOM took up to 223 millisecond while LINS only needed 25 millisecond. The results demonstrate the real-time capacity of LINS is much better than that of LIOM.

The most time-consuming parts in the LIO module of LIOM are local-map-constraint construction and batch optimization, in which it maintains a local map over multiple lidar scans and solves all relative states via MAP estimation. The main reason behind LINS's superior computational speed is that it uses Kalman filter rather than batch MAP, since Kalman filter implicitly reduces the dimension of the optimization problem by factoring the batch solution in time sequence and solving it in a recursive form \cite{barfoot2017state}. The other reason is that we only use point clouds from the previous lidar scan for matching. In this way, although the used point cloud is sparser than the local map built in LIOM, we can still achieve accurate results with the aids of the IMU.

\section{Conclusion}\label{Conclusion}
In this paper, we developed a lightweight lidar-inertial state estimator for robot navigation. Using an iterated ESKF with robocentric formulation, our algorithm is capable of providing real-time, long-term, robust, and high-precision ego-motion estimation under challenging environments. The proposed algorithm is verified in various scenarios including city, port, industrial park, forest, and indoor parking lot. Experimental results demonstrate that LINS outperforms the lidar-only methods and reaches comparable performance with the state-of-the-art lidar-inertial odometry with a much lower computational cost.








\balance
\bibliographystyle{bibtex/bst/IEEEtran}
\bibliography{bibtex/bib/IEEEabrv,bibtex/bib/LINS}

\begin{thebibliography}{10}
\providecommand{\url}[1]{#1}
\csname url@rmstyle\endcsname
\providecommand{\newblock}{\relax}
\providecommand{\bibinfo}[2]{#2}
\providecommand\BIBentrySTDinterwordspacing{\spaceskip=0pt\relax}
\providecommand\BIBentryALTinterwordstretchfactor{4}
\providecommand\BIBentryALTinterwordspacing{\spaceskip=\fontdimen2\font plus
\BIBentryALTinterwordstretchfactor\fontdimen3\font minus
  \fontdimen4\font\relax}
\providecommand\BIBforeignlanguage[2]{{%
\expandafter\ifx\csname l@#1\endcsname\relax
\typeout{** WARNING: IEEEtran.bst: No hyphenation pattern has been}%
\typeout{** loaded for the language `#1'. Using the pattern for}%
\typeout{** the default language instead.}%
\else
\language=\csname l@#1\endcsname
\fi
#2}}

\bibitem{velas2016collar}
M.~Velas, M.~Spanel, and A.~Herout, ``Collar line segments for fast odometry
  estimation from velodyne point clouds,'' in \emph{2016 IEEE International
  Conference on Robotics and Automation (ICRA)}.\hskip 1em plus 0.5em minus
  0.4em\relax IEEE, 2016, pp. 4486--4495.

\bibitem{barfoot2016into}
T.~D. Barfoot, C.~McManus, S.~Anderson, H.~Dong, E.~Beerepoot, C.~H. Tong,
  P.~Furgale, J.~D. Gammell, and J.~Enright, ``Into darkness: Visual navigation
  based on a lidar-intensity-image pipeline,'' in \emph{Robotics
  Research}.\hskip 1em plus 0.5em minus 0.4em\relax Springer, 2016, pp.
  487--504.

\bibitem{anderson2013ransac}
S.~Anderson and T.~D. Barfoot, ``Ransac for motion-distorted 3d visual
  sensors,'' in \emph{2013 IEEE/RSJ International Conference on Intelligent
  Robots and Systems}.\hskip 1em plus 0.5em minus 0.4em\relax IEEE, 2013, pp.
  2093--2099.

\bibitem{behley2018efficient}
J.~Behley and C.~Stachniss, ``Efficient surfel-based slam using 3d laser range
  data in urban environments.'' in \emph{Robotics: Science and Systems}, 2018.

\bibitem{ye2019tightly}
H.~Ye, Y.~Chen, and M.~Liu, ``Tightly coupled 3d lidar inertial odometry and
  mapping,'' in \emph{2019 IEEE International Conference on Robotics and
  Automation (ICRA)}.\hskip 1em plus 0.5em minus 0.4em\relax IEEE, 2019.

\bibitem{rusinkiewicz2001efficient}
S.~Rusinkiewicz and M.~Levoy, ``Efficient variants of the icp algorithm.'' in
  \emph{3dim}, vol.~1, 2001, pp. 145--152.

\bibitem{pomerleau2015review}
F.~C. Pomerleau, François and R.~Siegwart, ``A review of point cloud
  registration algorithms for mobile robotics,'' \emph{Foundations and Trends
  in Robotics}, vol.~4, no.~1, pp. 1--104, 2015.

\bibitem{zhang2014loam}
J.~Zhang and S.~Singh, ``Loam: Lidar odometry and mapping in real-time.'' in
  \emph{Robotics: Science and Systems}, vol.~2, 2014, p.~9.

\bibitem{shan2018lego}
T.~Shan and B.~Englot, ``Lego-loam: Lightweight and ground-optimized lidar
  odometry and mapping on variable terrain,'' in \emph{2018 IEEE/RSJ
  International Conference on Intelligent Robots and Systems (IROS)}.\hskip 1em
  plus 0.5em minus 0.4em\relax IEEE, 2018, pp. 4758--4765.

\bibitem{hess2016real}
W.~Hess, D.~Kohler, H.~Rapp, and D.~Andor, ``Real-time loop closure in 2d lidar
  slam,'' in \emph{2016 IEEE International Conference on Robotics and
  Automation (ICRA)}.\hskip 1em plus 0.5em minus 0.4em\relax IEEE, 2016, pp.
  1271--1278.

\bibitem{tang2015lidar}
J.~Tang, Y.~Chen, X.~Niu, L.~Wang, L.~Chen, J.~Liu, C.~Shi, and J.~Hyypp{\"a},
  ``Lidar scan matching aided inertial navigation system in gnss-denied
  environments,'' \emph{Sensors}, vol.~15, no.~7, pp. 16\,710--16\,728, 2015.

\bibitem{lynen2013robust}
S.~Lynen, M.~W. Achtelik, S.~Weiss, M.~Chli, and R.~Siegwart, ``A robust and
  modular multi-sensor fusion approach applied to mav navigation,'' in
  \emph{2013 IEEE/RSJ international conference on intelligent robots and
  systems}.\hskip 1em plus 0.5em minus 0.4em\relax IEEE, 2013, pp. 3923--3929.

\bibitem{zhen2017robust}
W.~Zhen, S.~Zeng, and S.~Soberer, ``Robust localization and localizability
  estimation with a rotating laser scanner,'' in \emph{2017 IEEE International
  Conference on Robotics and Automation (ICRA)}.\hskip 1em plus 0.5em minus
  0.4em\relax IEEE, 2017, pp. 6240--6245.

\bibitem{li2013real}
M.~Li, B.~H. Kim, and A.~I. Mourikis, ``Real-time motion tracking on a
  cellphone using inertial sensing and a rolling-shutter camera,'' in
  \emph{2013 IEEE International Conference on Robotics and Automation}.\hskip
  1em plus 0.5em minus 0.4em\relax IEEE, 2013, pp. 4712--4719.

\bibitem{huai2018robocentric}
Z.~Huai and G.~Huang, ``Robocentric visual-inertial odometry,'' in \emph{2018
  IEEE/RSJ International Conference on Intelligent Robots and Systems
  (IROS)}.\hskip 1em plus 0.5em minus 0.4em\relax IEEE, 2018, pp. 6319--6326.

\bibitem{qin2018vins}
T.~Qin, P.~Li, and S.~Shen, ``Vins-mono: A robust and versatile monocular
  visual-inertial state estimator,'' \emph{IEEE Transactions on Robotics},
  vol.~34, no.~4, pp. 1004--1020, 2018.

\bibitem{leutenegger2013keyframe}
S.~Leutenegger, P.~Furgale, V.~Rabaud, M.~Chli, K.~Konolige, and R.~Siegwart,
  ``Keyframe-based visual-inertial slam using nonlinear optimization,''
  \emph{Proceedings of Robotis Science and Systems (RSS) 2013}, 2013.

\bibitem{huang2011observability}
G.~P. Huang, N.~Trawny, A.~I. Mourikis, and S.~I. Roumeliotis,
  ``Observability-based consistent ekf estimators for multi-robot cooperative
  localization,'' \emph{Autonomous Robots}, vol.~30, no.~1, pp. 99--122, 2011.

\bibitem{park2018elastic}
C.~Park, P.~Moghadam, S.~Kim, A.~Elfes, C.~Fookes, and S.~Sridharan, ``Elastic
  lidar fusion: Dense map-centric continuous-time slam,'' in \emph{2018 IEEE
  International Conference on Robotics and Automation (ICRA)}.\hskip 1em plus
  0.5em minus 0.4em\relax IEEE, 2018, pp. 1206--1213.

\bibitem{geneva2018lips}
P.~Geneva, K.~Eckenhoff, Y.~Yang, and G.~Huang, ``Lips: Lidar-inertial 3d plane
  slam,'' in \emph{2018 IEEE/RSJ International Conference on Intelligent Robots
  and Systems (IROS)}.\hskip 1em plus 0.5em minus 0.4em\relax IEEE, 2018, pp.
  123--130.

\bibitem{forster2015imu}
C.~Forster, L.~Carlone, F.~Dellaert, and D.~Scaramuzza, ``Imu preintegration on
  manifold for efficient visual-inertial maximum-a-posteriori
  estimation.''\hskip 1em plus 0.5em minus 0.4em\relax Georgia Institute of
  Technology, 2015.

\bibitem{hesch2010laser}
J.~A. Hesch, F.~M. Mirzaei, G.~L. Mariottini, and S.~I. Roumeliotis, ``A
  laser-aided inertial navigation system (l-ins) for human localization in
  unknown indoor environments,'' in \emph{2010 IEEE International Conference on
  Robotics and Automation (ICRA)}.\hskip 1em plus 0.5em minus 0.4em\relax IEEE,
  2010, pp. 5376--5382.

\bibitem{huang2013quadratic}
G.~P. Huang, A.~I. Mourikis, and S.~I. Roumeliotis, ``A quadratic-complexity
  observability-constrained unscented kalman filter for slam,'' \emph{IEEE
  Transactions on Robotics}, vol.~29, no.~5, pp. 1226--1243, 2013.

\bibitem{huang2007convergence}
S.~Huang and G.~Dissanayake, ``Convergence and consistency analysis for
  extended kalman filter based slam,'' \emph{IEEE Transactions on robotics},
  vol.~23, no.~5, pp. 1036--1049, 2007.

\bibitem{barfoot2017state}
T.~D. Barfoot, \emph{State Estimation for Robotics}.\hskip 1em plus 0.5em minus
  0.4em\relax Cambridge University Press, 2017.

\bibitem{bell1993iterated}
B.~M. Bell and F.~W. Cathey, ``The iterated kalman filter update as a
  gauss-newton method,'' \emph{IEEE Transactions on Automatic Control},
  vol.~38, no.~2, pp. 294--297, 1993.

\bibitem{sola2017quaternion}
J.~Sola, ``Quaternion kinematics for the error-state kalman filter,''
  \emph{arXiv preprint arXiv:1711.02508}, 2017.

\bibitem{civera20101}
J.~Civera, O.~G. Grasa, A.~J. Davison, and J.~Montiel, ``1-point ransac for
  extended kalman filtering: Application to real-time structure from motion and
  visual odometry,'' \emph{Journal of Field Robotics}, vol.~27, no.~5, pp.
  609--631, 2010.

\bibitem{bloesch2017iterated}
M.~Bloesch, M.~Burri, S.~Omari, M.~Hutter, and R.~Siegwart, ``Iterated extended
  kalman filter based visual-inertial odometry using direct photometric
  feedback,'' \emph{The International Journal of Robotics Research}, vol.~36,
  no.~10, pp. 1053--1072, 2017.

\bibitem{madyastha2011extended}
V.~Madyastha, V.~Ravindra, S.~Mallikarjunan, and A.~Goyal, ``Extended kalman
  filter vs. error state kalman filter for aircraft attitude estimation,'' in
  \emph{AIAA Guidance, Navigation, and Control Conference}, 2011, p. 6615.

\bibitem{bloesch2016primer}
M.~Bloesch, H.~Sommer, T.~Laidlow, M.~Burri, G.~Nuetzi, P.~Fankhauser,
  D.~Bellicoso, C.~Gehring, S.~Leutenegger, M.~Hutter, \emph{et~al.}, ``A
  primer on the differential calculus of 3d orientations,'' \emph{arXiv
  preprint arXiv:1606.05285}, 2016.

\bibitem{shin2005estimation}
E.-H. Shin, ``Estimation techniques for low-cost inertial navigation,''
  \emph{UCGE report}, vol. 20219, 2005.

\bibitem{quigley2009ros}
M.~Quigley, K.~Conley, B.~Gerkey, J.~Faust, T.~Foote, J.~Leibs, R.~Wheeler, and
  A.~Y. Ng, ``Ros: an open-source robot operating system,'' in \emph{ICRA
  workshop on open source software}, vol.~3, no. 3.2.\hskip 1em plus 0.5em
  minus 0.4em\relax Kobe, Japan, 2009, p.~5.

\end{thebibliography}

\addtolength{\textheight}{-12cm}   

\end{document}